\documentclass{article}

\usepackage[preprint]{neurips_2026}

\usepackage[utf8]{inputenc} 
\usepackage[T1]{fontenc}    
\usepackage{hyperref}       
\usepackage{url}            
\usepackage{booktabs}       
\usepackage{amsfonts}       
\usepackage{nicefrac}       
\usepackage{microtype}      
\usepackage{xcolor}         
\usepackage{amsmath}
\usepackage{amssymb}
\usepackage{mathtools}
\usepackage{amsthm}
\usepackage{microtype}
\usepackage{graphicx}
\usepackage{subcaption}
\usepackage{booktabs} 
\usepackage{pgfplots}
\pgfplotsset{compat=1.17}
\usepackage{enumitem}
\usepackage{placeins}
\usepackage{natbib}
\usepackage{wrapfig}

\title{MetaCluster: Enabling Deep Compression of Kolmogorov-Arnold Networks}

\author{%
  Matthew Raffel\thanks{Equal contribution.} \qquad
  Adwaith Renjith\footnotemark[1] \qquad
  Lizhong Chen \\
  School of Electrical Engineering and Computer Science\\
  Oregon State University\\
  Corvallis, OR 97331 \\
  \texttt{\{raffelm, renjitha, chenliz\}@oregonstate.edu}
}

\begin{document}

\maketitle

\thispagestyle{plain}
\setcounter{footnote}{0}
\begin{abstract}
Kolmogorov-Arnold Networks (KANs) replace scalar weights with per-edge vectors of basis coefficients, thereby increasing expressivity and accuracy while also leading to a multiplicative increase in parameters and memory usage. We propose MetaCluster, a framework that makes KANs highly compressible without sacrificing accuracy. Specifically, a lightweight meta-learner, trained jointly with the KAN, maps low-dimensional embeddings to coefficient vectors, thereby constraining them to lie on a low-dimensional manifold amenable to clustering. We then run K-means in coefficient space and replace per-edge vectors with shared centroids. Afterward, the meta-learner can be discarded, and a brief fine-tuning of the centroid codebook recovers any residual accuracy loss. The resulting model stores only a small codebook and per-edge indices, exploiting the vector nature of KAN parameters to amortize storage across multiple coefficients. On MNIST, CIFAR-10, and CIFAR-100, across standard KANs and ConvKANs using multiple basis functions, MetaCluster achieves a reduction of up to $80\times$ in parameter storage, with minimal impact on accuracy. Similarly, in high-dimensional equation modeling tasks, MetaCluster achieves a $124.1\times$ parameter reduction while preserving baseline MSE.
\end{abstract}

\section{Introduction}
Kolmogorov–Arnold Networks (KANs) have recently emerged as a compelling alternative to multi-layer perceptrons (MLPs), delivering strong results in equation modeling and scientific machine learning, often achieving improved task performance with comparable or lower parameter counts in these settings \citep{liu2024kan, li2025u, coffman2025matrixkan, koenig2024kan}. Recently, KANs have also shown promise in computer vision \citep{yang2024kolmogorov, raffel2025flashkat}. However, unlike equation modeling, KANs at larger scales often incur substantial parameter overhead relative to MLPs \citep{yu2024kan}. This overhead stems from the per-edge degrees of freedom in KANs: each connection carries a vector of basis coefficients (e.g., B-spline weights) rather than a single scalar weight, resulting in a multiplicative increase in the number of parameters.

One approach that targets this multiplicative increase in parameters is weight sharing, which clusters parameters into a codebook that stores compact indices. Unfortunately, naively applying weight sharing to KANs is ineffective. Instead of clustering scalars (as in MLPs), we must cluster high-dimensional coefficient vectors in KANs.  In such high-dimensional spaces, absolute distances increase and become more concentrated (so that the nearest and farthest points become similar), causing clustering methods to struggle to form tight clusters \citep{beyer1999nearest,donoho2000high}.

We address this challenge with MetaCluster\footnote{Our code is publicly available at \url{https://github.com/OSU-STARLAB/MetaCluster}.}, a three-stage compression framework that merges meta-learning with weight sharing. First, a small meta-learner maps low-dimensional embeddings to per-edge coefficient vectors, thereby constraining KAN activations to a low-dimensional manifold during training on the task loss. This manifold shaping makes the coefficient vectors highly clusterable. Second, we run K-means on the generated coefficients, replacing per-edge weights with codebook centroids indexed with compact codes.  Finally, we discard the meta-learner and embeddings and lightly fine-tune the centroids to recover any lost accuracy. Since each centroid stores an entire coefficient vector, the codebook amortizes over many scalars, yielding a much higher compression factor for KANs than for MLPs at the same number of clusters.

We validate MetaCluster on two model families consisting of a fully-connected KAN \citep{liu2024kan} and a convolutional KAN (ConvKAN) \citep{bodner2024convolutional, drokin2024kolmogorov}. For each of these models, we test the efficacy of using B-Splines, radial basis functions (RBFs), and Gram polynomials as the bases \citep{liu2024kan, li2024kolmogorov,ss2024chebyshev}. To further verify the robustness of our approach, we validate it across MNIST \citep{6296535}, CIFAR-10, CIFAR-100 \citep{Krizhevsky09learningmultiple}, and equation modeling. From our experiments, we find that MetaCluster achieves up to $80\times$ and $124.1\times$ reductions in parameter storage for image classification and equation modeling, respectively, relative to the uncompressed KAN while closely matching original performance.  We also find it is robust across various architectures and datasets, and ablations confirm that enforcing a low-dimensional manifold is key to high-quality clustering. 

The main contributions of this paper are:
\begin{enumerate}[noitemsep, topsep=0pt]
    \item We identify the potential of weight-sharing for reducing the KANs memory footprint and, to our knowledge, provide the first effective weight-sharing method tailored to KANs.
    \item We propose a meta-learning approach that shapes per-edge KAN coefficients to lie on a low-dimensional manifold, enabling effective clustering in high dimensions.
    \item We provide extensive image classification and equation modeling experiments demonstrating up to $80\times$ and $124.1\times$ memory reduction, respectively, while largely preserving accuracy, along with extensive ablations.
\end{enumerate}

\section{Preliminaries and Motivation}
\subsection{Kolmogorov-Arnold Networks}
KANs have gained attention as an alternative to conventional MLPs \citep{liu2024kan}. Their design is motivated by the Kolmogorov–Arnold representation theorem, which guarantees that any continuous multivariate function on a bounded domain, $f:[0,1]^n\rightarrow\mathbb{R}$, can be expressed as a finite sum of compositions of univariate continuous functions, $\phi_{q,p}:[0,1]\rightarrow \mathbb{R}$ and $\phi_q:\mathbb{R} \rightarrow \mathbb{R}$ such that

\begin{equation}
    \label{eq:KAN}
    f(\mathbf{x}) = f(x_1, ..., x_n) = \sum^{2n+1}_{q=1}\phi_q(\sum^n_{p=1}\phi_{q,p}(x_p)).
\end{equation}

While Equation \ref{eq:KAN} captures the classical theoretical form, \citet{liu2024kan} demonstrates a practical generalization that permits arbitrary width and depth tailored to the task. We can express such a formulation with $L$ layers as
\begin{align}
    \label{eq:KANDeep}
    f(\mathbf{x})&=(\Phi_L \circ \Phi_{L-1} \circ ... \circ \Phi_1)\mathbf{x}, \nonumber \\  
    \Phi_l &=
\begin{bmatrix}
\phi_{l,1,1}(\,\cdot\,) & \cdots & \phi_{l,1,n_l}(\,\cdot\,) \\
\vdots               & \ddots & \vdots               \\
\phi_{l,n_{l+1},1}(\,\cdot\,) & \cdots & \phi_{l, n_{l+1},n_l}(\,\cdot\,)
\end{bmatrix}
\end{align}
In Equation \ref{eq:KANDeep}, we let $n_l$ denote the number of inputs to layer $l$, with inputs $x_{l, i}$. Each edge from input $i$ to output $j$ in layer $l$ has a learnable univariate activation $\phi_{l,j,i}(\cdot)$, for $i\in[1,n_l]$ and $j\in[1,n_{l+1}]$.

The most common choice for implementing $\phi_{l,j,i}(\cdot)$ has been through a weighted summation of basis functions $B_i(\cdot)$ parameterized with $\mathbf{w}=[w_1, ..., w_{|\mathbf{w}|}]$ as shown with,
\begin{equation}
	\phi_{l,j,i}(\cdot) =  \sum_{i=1}^{|\mathbf{w}|} w_i B_i(\cdot).
    \label{eq:bspline}
\end{equation}

From its increased representation power compared to the MLP, the KAN has offered impressive results on equation modeling tasks for scientific applications \citep{li2024kolmogorov, li2025u, coffman2025matrixkan, koenig2024kan} and has even started to extend its reach into computer vision \citep{raffel2025flashkat, yang2024kolmogorov}. Despite these gains, widespread adoption in modern large-scale architectures has been hindered by memory inefficiency \citep{yu2024kan}.  The root cause is structural: each KAN edge carries a vector of basis coefficients (e.g., B-spline weights), whereas an MLP edge carries a single scalar. Ignoring biases, an arbitrary MLP with $L$ layers will possess $\sum_{l=0}^{L-1}(n_{l}\times n_{l+1})$ parameters, whereas a KAN will possess $\sum_{l=0}^{L-1}(n_{l}\times n_{l+1})\times(|\mathbf{w}|)$ parameters.  Thus, for identical topologies, a KAN is approximately $|\mathbf{w}|$ times larger than its MLP counterpart. 

\subsection{KAN Weight Sharing Motivations and Challenges}
These observations motivate a compression strategy that targets the dimensionality of the vector of basis coefficients. Weight sharing is one method that reduces dimensionality directly by clustering parameters into a small codebook and storing compact indices. In its classical form for MLPs, one applies K-means to the set of scalar weights $\mathbf{W}={w_1,w_2,\dots,w_n}$, obtaining centroids $\mathbf{C}={c_1,c_2,\dots,c_k}$ and assignments that minimize within-cluster sum of squares \citep{han2015deep}:
\begin{equation}\label{eq:K-means}
\arg\min_{\mathcal C}
\sum_{i=1}^{k}
\sum_{w\in C_i}
(w - c_i)^2
\end{equation}

Weight sharing has repeatedly been shown to preserve accuracy while substantially reducing the number of parameters \citep{han2015deep, cho2021dkm}. In MLPs, however, their effectiveness is constrained by the need to store, for every weight, an index that maps back to a codebook centroid; an overhead that limits the ultimate compression. Under a simple model with $n$ weights, a codebook of $k$ centroids stored at $b$ bits per scalar, the achievable compression factor is
\begin{equation}
    r=\frac{nb}{n\log_2(k)+kb}
    \label{eq:MLPcompression}
\end{equation}
The dominant term in the denominator, $n \log_2 k$, represents the per-weight index cost required to record the mapping from each original weight to its corresponding centroid. When extending weight sharing to KANs, the index mapping is still required, but its relative cost can be significantly reduced because we cluster $|\mathbf{w}|$-dimensional coefficient vectors per edge.  Thus, each centroid stores $|\mathbf{w}|$ scalars, amortizing the codebook over many parameters while the index cost remains $n\log_2 k$. We detail this amortization and its impact on compression in Section \ref{sec:storage}.

While weight sharing is promising for substantially reducing storage in KANs, directly applying it to KANs for model compression is nontrivial.  For instance, since each edge on a KAN is defined by $|\mathbf{w}|$ weights rather than $1$ as in the MLP, the dimensionality of the points we must cluster is far greater.  In increasing the dimensionality, the vector space becomes sparser and points become more equally spaced apart, making clustering points more difficult \citep{beyer1999nearest,donoho2000high}. Therefore, to effectively cluster KAN activations, we require a fundamentally different method to transform the high-dimensional vector into a more manageable space. 

\section{Methods}
\label{sec:methods}
This section presents MetaCluster, a three-stage framework that makes per-edge KAN activation coefficients amenable to clustering and compression. First, we train a single meta-learner that maps compact embeddings to full activation-coefficient vectors, thereby forcing them onto a task-aligned low-dimensional manifold. Next, we apply K-means to those vectors and record, for each edge, the centroid to which it belongs in that layer. Finally, we discard the meta-learner and embeddings, replace per-edge weights with their assigned centroids via a lookup table, and briefly fine-tune the network to recover any lost accuracy.

\subsection{Manifold Learning through a Meta-learner}
\label{sec:manifold}
Clustering the weights of a KAN is challenging due to the high dimensionality of each weight vector. To address this, we introduce a single meta-learner, $M_\theta$, which maps a lower-dimensional embedding $\mathbf{z}_i \in \mathbb{R}^{d_{\text{emb}}}$ to the full KAN weights $\mathbf{w}_i \in \mathbb{R}^{|\mathbf{w}|}$, constraining them to lie on a low-dimensional manifold. Formally, the mapping is defined as:

\begin{equation}
M_\theta(\mathbf{z}_i) = \mathbf{W}_2 \sigma(\mathbf{W}_1 \mathbf{z}_i + b_1) + b_2 = \mathbf{w}_i,
\end{equation}

where $W_1 \in \mathbb{R}^{d_{\text{hidden}} \times d_{\text{emb}}}$, $W_2 \in \mathbb{R}^{|\mathbf{w}| \times d_{\text{hidden}}}$, and $\sigma(\cdot)$ is a ReLU activation. The meta-learner is trained jointly with the KAN using standard backpropagation for $\alpha$ epochs, so that the generated weights $\mathbf{w_i}$ both lie on a meaningful manifold and optimize the task-specific loss.

\begin{figure*}[t]
  \centering
  \begin{subfigure}{0.32\textwidth}
    \centering
    \includegraphics[width=\linewidth]{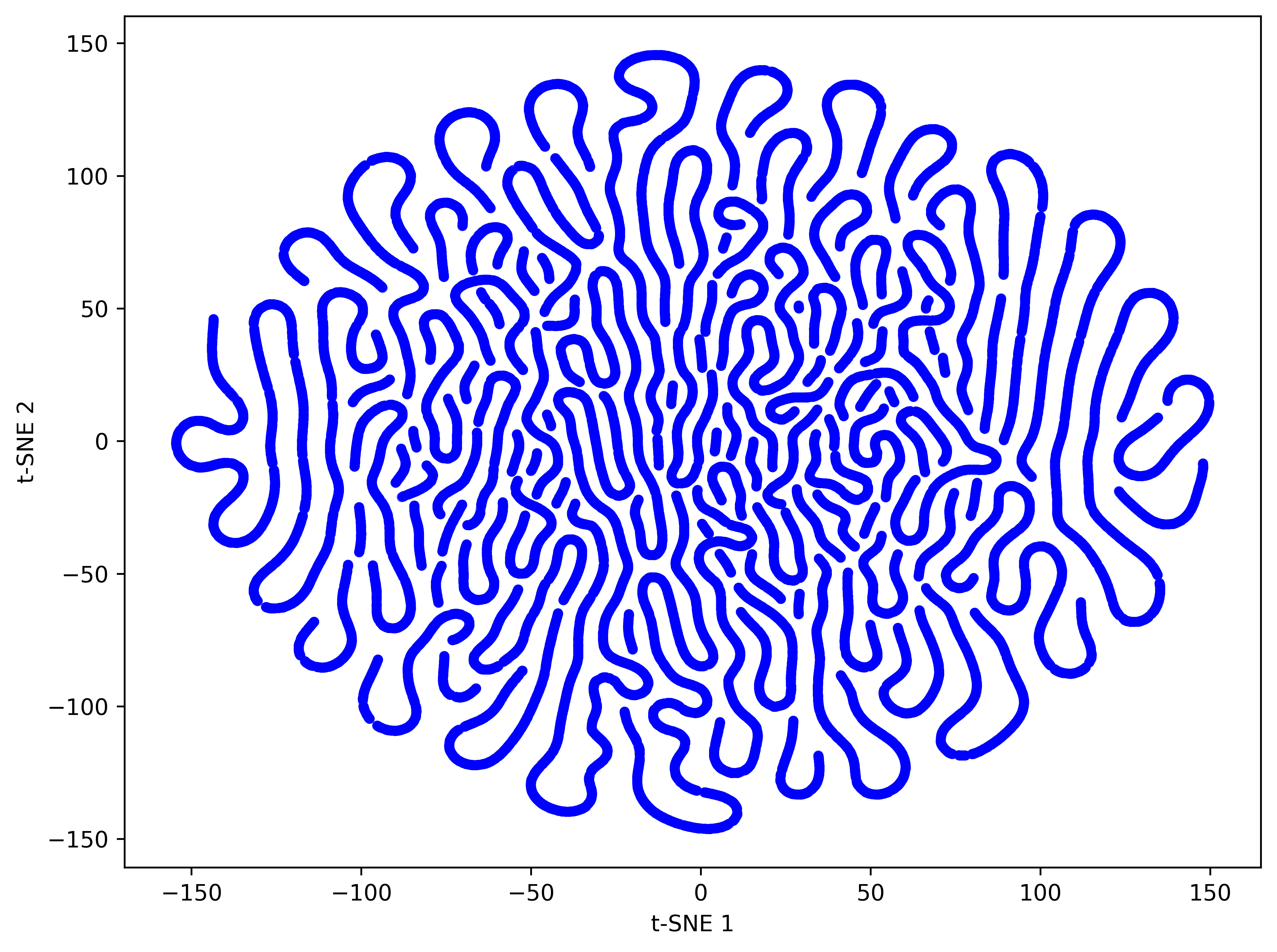}
    \caption{Meta-learner when $d_{emb}=1$}
    \label{fig:Meta1TSNE}
  \end{subfigure}\hfill
  \begin{subfigure}{0.32\textwidth}
    \centering
    \includegraphics[width=\linewidth]{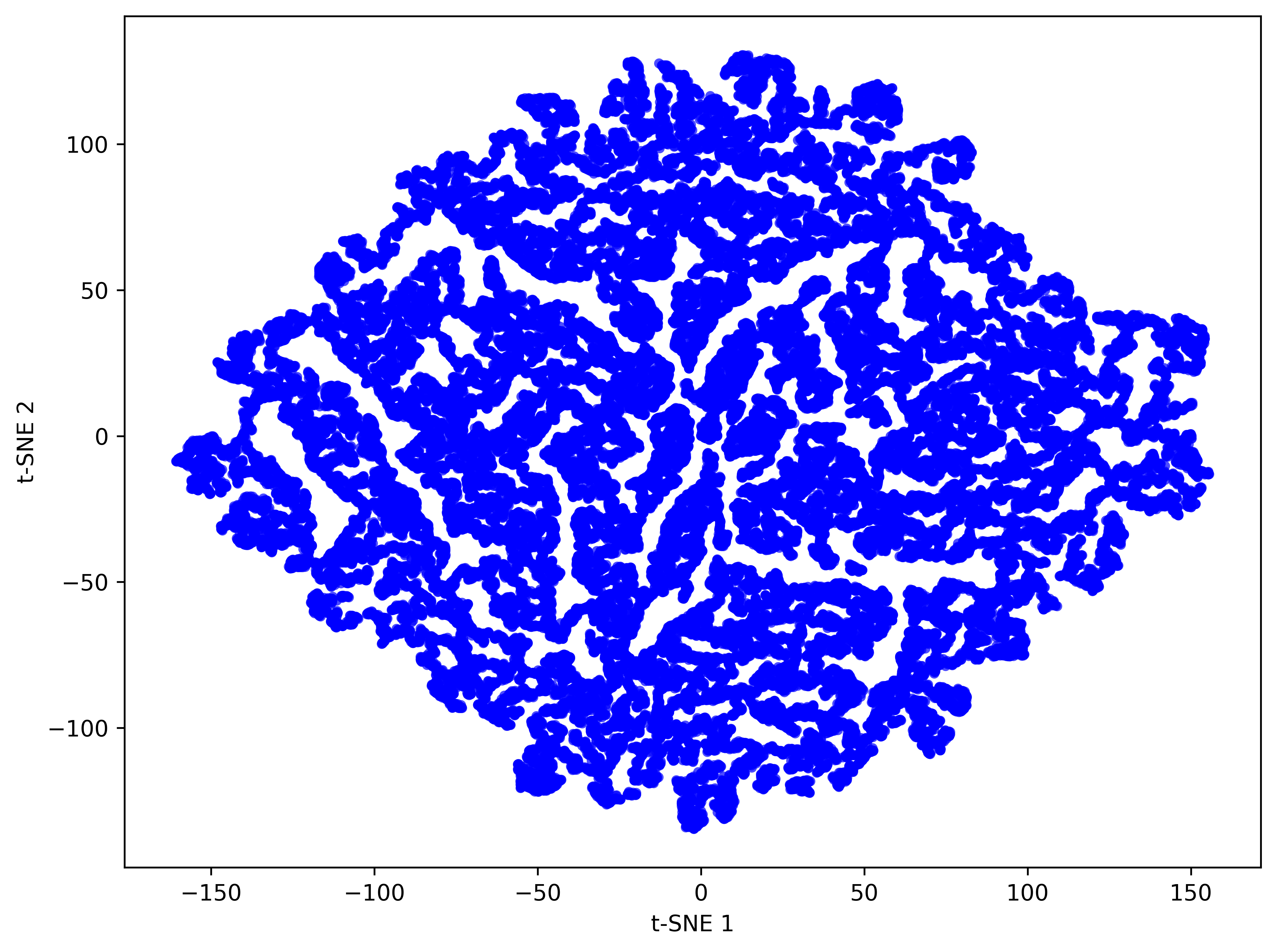}
    \caption{Meta-learner when $d_{emb}=2$}
    \label{fig:Meta2TSNE}
  \end{subfigure}\hfill
  \begin{subfigure}{0.32\textwidth}
    \centering
    \includegraphics[width=\linewidth]{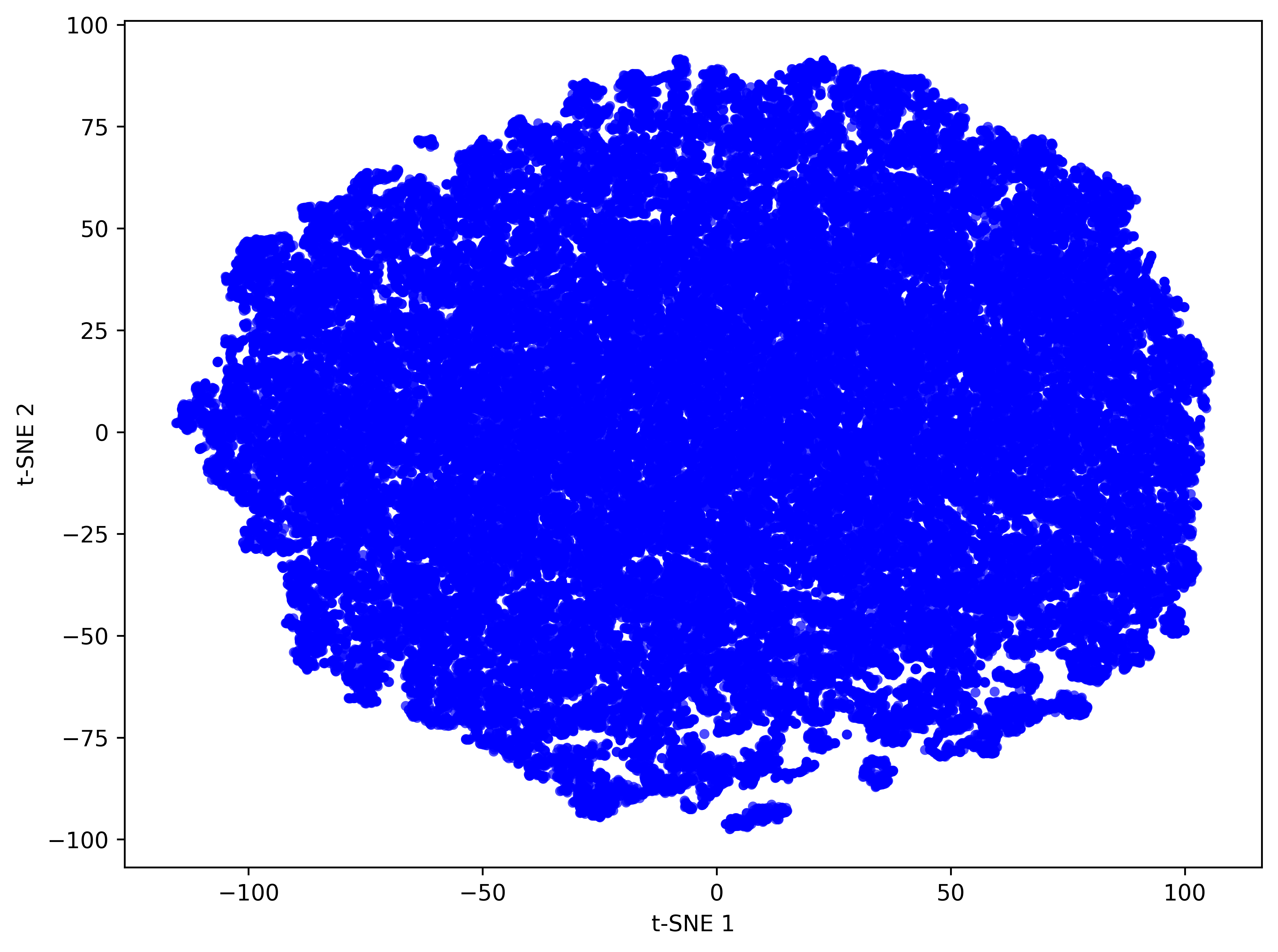}
    \caption{KAN}
    \label{fig:KANTSNE}
  \end{subfigure}

  \caption{T-SNE visualization of KAN activation weights with and without the meta-learner $M_\theta$. (a) With a 1D embedding, the MetaKAN learns a nearly one-dimensional manifold, (b) with a 2D embedding, it learns a structured two-dimensional manifold, while (c) the baseline KAN without a meta-learner fails to organize the weights into a coherent low-dimensional structure. 
  }
  \label{fig:tsne}
\end{figure*}

To demonstrate the manifold-shaping effect of the meta-learner, we train a two-layer KAN with a B-spline basis on CIFAR-10 \citep{Krizhevsky09learningmultiple} for one epoch under three settings: (i) a meta-learner with $d_{\text{emb}}=1$, (ii) a meta-learner with $d_{\text{emb}}=2$, and (iii) a baseline KAN without a meta-learner. We collect all per-edge activation coefficient vectors ${\mathbf{w}_i}$ and visualize them with t-SNE. Each point corresponds to $\mathbf{w}_i \in \mathbb{R}^{|\mathbf{w}|}$, where $|\mathbf{w}|=9$. As shown in Figure \ref{fig:Meta1TSNE}, when $d_{\text{emb}}=1$ the model organizes coefficients along an effectively one-dimensional manifold. Increasing to $d_{\text{emb}}=2$ produces the well-structured two-dimensional sheet shown in Figure \ref{fig:Meta2TSNE}. 
In contrast, the baseline KAN provided in Figure \ref{fig:KANTSNE} yields a diffuse cloud without coherent low-dimensional structure, which hampers downstream clustering. This qualitative observation is quantitatively supported by our Principal Component Analysis and cluster validation metrics in Appendix \ref{app:pca}, which confirm that the meta-learner effectively restricts the variance of the coefficient space to the target dimensionality \citep{MACKIEWICZ1993303}. These results substantiate our premise: shaping coefficient vectors via a low-dimensional embedding makes them markedly more clusterable, a property we later translate into higher compression at fixed accuracy, as shown in Section \ref{sec:ablation}.

\subsection{KAN  Activation Compression with MetaClustering}

We will now outline our method for clustering the weights $M_{\theta}(\mathbf{z}_i)$ using K-means.  Unlike Equation \ref{eq:K-means}, the weights $M_{\theta}(\mathbf{z}_i)$ lie in $\mathbb{R}^{|\mathbf{w}|}$ rather than $\mathbb{R}$ as with the scalar weights $w_i$.  
Accordingly, each centroid $\mathbf{c}_i$ also resides in $\mathbb{R}^{|\mathbf{w}|}$.  
Therefore, our K-means objective must be adapted as follows: 
\begin{equation}\label{eq:K-means_meta}
\arg\min_{\mathcal C}
\sum_{i=1}^{k}
\sum_{M_{\theta}(\mathbf{z})\in C_i}
\bigl\|M_{\theta}(\mathbf{z}) - \mathbf{c}_i\bigr\|_2^2,
\end{equation}
where $\mathbf{c}_i \in \mathbb{R}^{|\mathbf{w}|}$ denotes the centroid of cluster $C_i$.  

The centroids are updated iteratively according to the standard K-means update rule:
\begin{equation}
\mathbf{c}_i = \frac{1}{|C_i|} \sum_{M_{\theta}(\mathbf{z}_j) \in C_i} M_{\theta}(\mathbf{z}_j),
\end{equation}
ensuring that each centroid represents the mean of the weight vectors assigned to its cluster.  

To record assignments, we define an index mapping vector $I \in \{1,2,\dots,k\}^N$, where the $j$-th entry indicates the cluster index of $M_{\theta}(\mathbf{z}_j)$.  
Formally,
\begin{equation}
I_j = \arg\min_{i \in \{1,\dots,k\}} 
\bigl\|M_{\theta}(\mathbf{z}_j) - \mathbf{c}_i \bigr\|_2^2,
\end{equation}
so that the mapping is expressed as $M_{\theta}(\mathbf{z}_j) \;\mapsto\; \mathbf{c}_{I_j}$. We perform this clustering independently per layer, which allows the codebook to adapt to the distinct dynamic weight ranges across network depth.

\subsection{Accuracy Recovery with Brief Fine-tuning}
Once we have determined the mapping $I$ and the centroids $C$, we can remove the meta-learner $M_\theta$ and the embeddings $\mathbf{z}$ from the network as they are no longer needed (saving more memory in addition to compression). 
Since the clustered weights are an approximation of the original weights, we can recover any lost performance by fine-tuning the network.  The fine-tuning process consists of an identical procedure to the original training, in which we optimized the centroid codebook for the task-specific loss for $\beta$ epochs, where $\beta<<\alpha$ (the number of training epochs).

\subsection{Storage Analysis}
\label{sec:storage}

The compression factor relative to KAN is given by:
\begin{equation}
r = \frac{n |\mathbf{w}| b}{n\log_2(k) + |\mathbf{w}| k b} = \frac{nb}{n\log_2(k) / |\mathbf{w}| + kb}
\label{KANcompression}
\end{equation}
As in Equation \ref{eq:MLPcompression}, $n$ denotes the number of connections, $k$ the number of centroids, and $b$ the number of bits used to represent each edge. Notably, the term $|\mathbf{w}|$ in the denominator reduces the relative contribution of $n\log_2(k)$ to the overall storage. Intuitively, this happens because each centroid stores more weight information ($|\mathbf{w}|$ entries per centroid), so the overhead of indexing and storing the centroids becomes proportionally smaller. As a result, KAN benefits more from the compression scheme than MLP.

\section{Experiments} 
\label{sec:experiments}
We provide extensive experiments to showcase the efficacy of MetaCluster.  In Section \ref{sec:image_results}, we present the main results of the fully connected and convolutional architectures on MNIST, CIFAR10, and CIFAR100 \citep{6296535, Krizhevsky09learningmultiple}.  Next, in Section \ref{sec:ablation}, we provide the ablation study for our approach, covering the influence of the KAN meta-learner embedding size, basis coefficient vector sizes, and the cluster count on CIFAR10 \citep{Krizhevsky09learningmultiple}. In Section \ref{sec: computation} we report the computational cost incurred from MetaCluster. Finally, in Section \ref{sec: equation_modeling}, we provide high-dimensional equation modeling results with MetaCluster.  Across all experiments, our reported memory measurements reflect the full deployable footprint: per-edge indices are stored as compact integer tensors (4 bits for $k=16$ and 8 bits for $k=256$), centroid codebooks are stored in FP32, and basis evaluation grids are generated on the fly.

\begin{table*}[ht]
	\setlength{\tabcolsep}{3pt} 
	\caption{Classification accuracy and memory (KB) comparison of a fully-connected network.}
	\label{tab:fc_cv}
	\centering
	\begin{scriptsize}
		\begin{sc}
			\begin{tabular}{lccc|ccc|ccc}
				\toprule
				\textbf{Model} & \multicolumn{2}{c}{\textbf{MNIST}} & & \multicolumn{2}{c}{\textbf{CIFAR-10}} & & \multicolumn{2}{c}{\textbf{CIFAR-100}} & \\
				\cmidrule(lr){2-3} \cmidrule(lr){5-6} \cmidrule(lr){8-9}
				& Memory & Acc. & & Memory & Acc. & & Memory & Acc. \\
				\midrule
                KAN & 1,031.44 & $95.37 \pm 0.11$ & & 3,064.95 & $47.52 \pm 0.16$ & & 3,177.45 & $17.80 \pm 0.10$ \\
                MetaKAN & 141.32 & $96.36 \pm 0.09$ & & 410.82 & $46.28 \pm 0.39$ & & 422.08 & $20.17 \pm 0.24$ \\
                ClusterKAN & 13.84 & $61.37 \pm 2.77$ & & 38.34 & $27.92 \pm 2.88$ & & 39.75 & $6.20 \pm 0.31$ \\
                \hspace{2em}+ fine-tune & 13.84 & $92.66 \pm 0.08$ & & 38.34 & $43.71 \pm 0.27$ & & 39.75 & $11.04 \pm 0.64$ \\
				MetaClusterKAN & 13.84 & $96.06 \pm 0.10 $ & & 38.34 & $45.47 \pm 0.44$ & & 39.75 & $18.97 \pm 0.15$ \\
                \hspace{2em}+ fine-tune & 13.84 & $96.41 \pm 0.10$ & & 38.34 & $45.95 \pm 0.33$ & & 39.75 & $19.16 \pm 0.17$ \\
                \midrule
                FastKAN & 900.56 & $96.78 \pm 0.08$ & & 2,676.83 & $49.10 \pm 0.29$ & & 2,778.08 & $20.56 \pm 0.09$ \\
                MetaFastKAN & 108.09 & $95.51 \pm 0.12$ & & 316.35 & $47.40 \pm 0.46$ & & 327.60 & $19.14 \pm 0.40$ \\
                ClusterFastKAN & 20.54 & $32.34 \pm 4.16$ & & 57.30 & $24.35 \pm 0.83$ & & 58.71 & $6.72 \pm 0.41$ \\
                \hspace{2em}+ fine-tune & 20.54 & $86.24 \pm 1.08$ & & 57.30 & $38.89 \pm 0.35$ & & 58.71 & $11.35 \pm 0.34$ \\
				MetaClusterFastKAN & 20.54 & $94.87 \pm 0.29$ & & 57.30 & $46.96 \pm 0.44$ & & 58.71 & $18.66 \pm 0.48$ \\
                \hspace{2em}+ fine-tune & 20.54 & $95.59 \pm 0.18$ & & 57.30 & $47.60 \pm 0.37$ & & 58.71 & $19.05 \pm 0.47$ \\
                \midrule
                GramKAN & 497.46 & $96.76 \pm 0.08$ & & 1,477.48 & $49.57 \pm 0.43$ & & 1,534.43 & $21.89 \pm 0.09$ \\
                MetaGramKAN & 101.49 & $95.48 \pm 0.12$ & & 297.49 & $48.24 \pm 0.19$& & 309.45 & $19.11 \pm 0.24$\\
				ClusterGramKAN & 13.96 & $68.72 \pm 1.65$ & & 38.46 & $31.13 \pm 0.64$ & & 40.58 & $6.34 \pm 0.43$ \\
                \hspace{2em}+ fine-tune & 13.96 & $95.40 \pm 0.15$ & & 38.46 & $45.81 \pm 0.66$ & & 40.58 & $13.98 \pm 0.44$ \\
				MetaClusterGramKAN & 14.15 & $93.80 \pm 0.72$ & & 38.65 & $47.66 \pm 0.25$ & & 40.76 & $18.18 \pm 0.20$\\
                \hspace{2em}+ fine-tune & 14.15 & $95.32 \pm 0.12$ & & 38.65 & $48.47 \pm 0.30$ & & 40.76 & $18.96 \pm 0.25$\\
				\bottomrule
			\end{tabular}
		\end{sc}
	\end{scriptsize}
\end{table*}
\subsection{Image Classification Results}
\label{sec:image_results}
We compare our approach across 24 model schemes.  The base model for each scheme consists of KAN (B-spline basis) \citep{liu2024kan}, FastKAN (RBF basis) \citep{li2024kolmogorov}, and GramKAN (Gram polynomial basis)\citep{drokin2024kolmogorov}.  In our naming conventions, we include \textit{Cluster} to designate a clustered model, \textit{Meta} to designate a model using a meta-learner, a design that originates from \citet{zhao2025improving}, and \textit{Conv} to designate a model with a convolutional architecture \citep{bodner2024convolutional, drokin2024kolmogorov}. This choice balances edge density against index overhead, where a compact 4-bit codebook ($k=16$) suffices for smaller fully-connected networks, while denser convolutional models require an 8-bit codebook ($k=256$) to maintain capacity. Each result is reported as an average of 5 runs with standard error bars.  A full description of our hyperparameters is given in Appendices \ref{sec:fc_architecture} and \ref{sec:conv_architecture}.  


\subsubsection{Fully-Connected Image Classification}
\label{sec:full_image_results}
 
 Table \ref{tab:fc_cv} demonstrates the high classification accuracy and compression factor of fully-connected MetaCluster models across all datasets and basis functions. For example, on Cifar-10, MetaClusterKAN achieves a compression factor of up to $10.7\times$ (from 410.82KB to 38.34KB) compared with MetaKAN and up to $79.9\times$ (from 3064.95 KB to 38.34KB) compared with KAN.  Furthermore, it achieves this compression factor while matching the classification accuracy of MetaKAN.  Although the Meta variants can achieve high accuracy without fine-tuning, this is not the case for the non-Meta variants.  For example, in the case of GramKAN, it experiences a decrease in classification accuracy of up to $3.45\times$ (from 21.89\% to 6.34\%), which, when recovered with fine-tuning, remains a $1.57\times$ decrease in accuracy (from 21.89\% to 13.98 \%).  Such results demonstrate the importance of leveraging a meta-learner to learn a set of activations that lie in a lower-dimensional subspace to improve classification accuracy after clustering in fully-connected models. We also compare MetaCluster models against GPTVQ \citep{van2024gptvq}, a popular vector quantization technique that performs hessian-aware weight sharing.  From our experiments, we find that on Cifar-10, GPTVQ achieves only 31.83\% classification accuracy, compared to MetaClusterKAN, which achieves 43.47\%. Such results persist across all the datasets and all the models as we show in Appendix \ref{app:gptvq}.

\subsubsection{Convolutional Image Classification}
\begin{table*}[ht]
  \setlength{\tabcolsep}{3pt} 
  \caption{Classification accuracy and memory (KB) comparison of a convolutional network.}
  \label{tab:conv_cv}
  \centering
  \begin{scriptsize}
    \begin{sc}
      \begin{tabular}{lccc|ccc|ccc}
        \toprule
        \textbf{Model}
          & \multicolumn{2}{c}{\textbf{MNIST}}  & 
          & \multicolumn{2}{c}{\textbf{CIFAR-10}} &
          & \multicolumn{2}{c}{\textbf{CIFAR-100}} &      \\
        \cmidrule(lr){2-3} \cmidrule(lr){5-6} \cmidrule(lr){8-9}
          & Memory & Acc.\  & & Memory & Acc.\  & & Memory & Acc.\  \\
        \midrule
        KANConv
          & 13,634.00 & \(99.46 \pm 0.02\)   &
          & 13,654.27 & \(70.66 \pm 2.89\)   &
          & 13,744.62 & \(23.99 \pm 0.98\)   \\

        MetaKANConv
          & 3,048.40 & \(99.32 \pm 0.05\)   &
          & 3,052.90 & \(68.36 \pm 2.84\)   &
          & 3,143.25 & \(35.60 \pm 1.78\)   \\

        ClusterKANConv
          & 429.95 & \(92.16 \pm 1.24\)    &
          & 430.51 & \(33.26 \pm 1.30\)    &
          & 520.87 & \(10.81 \pm 0.54\)   \\

        \hspace{2em}+ fine-tune
          & 429.95 & \(98.89 \pm 0.09\)    &
          & 430.51 & \(61.80 \pm 0.34\)    &
          & 520.87 & \(22.65 \pm 0.81\)   \\

        MetaClusterKANConv
          & 434.77 & \(99.29 \pm 0.06\)    &
          & 435.34 & \(62.48 \pm 2.43\)    &
          & 525.70 & \(29.62 \pm 1.65\)   \\

        \hspace{2em}+ fine-tune
          & 434.77 & \(99.35 \pm 0.05\)    &
          & 435.34 & \(66.85 \pm 3.11\)    &
          & 525.70 & \(35.11 \pm 1.79\)   \\

        \midrule

        FastKANConv
          & 13,632.09 & \(99.11 \pm 0.09\)  &
          & 13,652.37 & \(76.30 \pm 0.11\)  &
          & 13,742.72 & \(46.36 \pm 0.38\)   \\

        MetaFastKANConv
          & 3,041.57 & \(99.00 \pm 0.11\)   &
          & 3,046.08 & \(69.06 \pm 1.15\)   &
          & 3,136.43 & \(39.60 \pm 1.13\)   \\

        ClusterFastKANConv
          & 428.03 & \(62.61 \pm 2.70\)    &
          & 428.61 & \(26.01 \pm 3.55\)    &
          & 518.97 & \(5.16 \pm 0.31\)    \\

        \hspace{2em}+ fine-tune
          & 428.03 & \(96.10 \pm 0.09\)    &
          & 428.61 & \(57.45 \pm 1.53\)    &
          & 518.97 & \(20.15 \pm 0.70\)   \\

        MetaClusterFastKANConv
          & 427.97 & \(99.01 \pm 0.11\)    &
          & 428.55 & \(68.72 \pm 1.06\)    &
          & 518.91 & \(38.75 \pm 1.04\)   \\

        \hspace{2em}+ fine-tune
          & 427.97 & \(99.08 \pm 0.06\)    &
          & 428.55 & \(69.10 \pm 1.10\)    &
          & 518.91 & \(39.26 \pm 1.10\)   \\

        \midrule

        GramKANConv
          & 7,586.47 & \(99.40 \pm 0.06\)  &
          & 7,597.72 & \(78.56 \pm 1.14\)  &
          & 7,688.07 & \(48.25 \pm 0.59\)   \\

        MetaGramKANConv
          & 3,047.93 & \(99.44 \pm 0.05\)  &
          & 3,052.43 & \(80.91 \pm 0.41\)  &
          & 3,142.79 & \(52.07 \pm 0.63\)   \\

        ClusterGramKANConv
          & 418.93 & \(61.17 \pm 9.40\)    &
          & 419.49 & \(12.08 \pm 0.66\)    &
          & 509.85 & \(1.77 \pm 0.29\)    \\

        \hspace{2em}+ fine-tune
          & 418.93 & \(99.27 \pm 0.08\)    &
          & 419.49 & \(75.74 \pm 0.63\)    &
          & 509.85 & \(44.01 \pm 0.42\)   \\

        MetaClusterGramKANConv
          & 418.81 & \(99.41 \pm 0.06\)    &
          & 419.38 & \(79.25 \pm 0.59\)    &
          & 509.73 & \(49.70 \pm 0.51\)   \\

        \hspace{2em}+ fine-tune
          & 418.81 & \(99.47 \pm 0.02\)    &
          & 419.38 & \(80.85 \pm 0.41\)    &
          & 509.73 & \(51.20 \pm 0.67\)   \\
        \bottomrule
      \end{tabular}
    \end{sc}
  \end{scriptsize}
\end{table*}
\begin{table*}[t]
  \setlength{\tabcolsep}{2pt}
  \caption{Impact of basis coefficient count on memory (KB) footprint  and classification accuracy. 
  }
  \label{tab:grid_size}
  \centering
  \begin{scriptsize}
    \begin{sc}
      \begin{tabular}{l|rr rrr|rr rrr}
        \toprule
           & \multicolumn{2}{c}{MetaFastKAN}
           & \multicolumn{3}{c}{MetaClusterFastKAN}
           & \multicolumn{2}{c}{MetaFastKANConv}
           & \multicolumn{3}{c}{MetaClusterFastKANConv} \\
        \cmidrule(lr){2-3}
        \cmidrule(lr){4-6}
        \cmidrule(lr){7-8}
        \cmidrule(lr){9-11}
         
    \multicolumn{1}{c}{Coeff}
      & \multicolumn{1}{c}{Memory} & \multicolumn{1}{c}{Acc\%}
      & \multicolumn{1}{c}{Memory} & \multicolumn{1}{c}{Acc\%} & \multicolumn{1}{c}{\%Chg}
      & \multicolumn{1}{c}{Memory} & \multicolumn{1}{c}{Acc\%}
      & \multicolumn{1}{c}{Memory} & \multicolumn{1}{c}{Acc\%} & \multicolumn{1}{c}{\%Chg} \\
        \midrule
        5  & 316.35 & $47.40 \pm 0.46$
           & 57.30 & $47.60 \pm 0.37$ & 0.42
           & 3,046.08 & $69.97 \pm 0.62$
           & 428.55 & $69.73 \pm 0.63$ & -0.34 \\

        8  & 316.76 & $46.29 \pm 0.19$
           & 57.68 & $46.30 \pm 0.24$ & 0.02
           & 3,046.52 & $68.89 \pm 0.63$
           & 440.61 & $68.92 \pm 0.60$ & 0.05 \\

        10 & 317.03 & $46.63 \pm 0.23$
           & 57.93 & $47.08 \pm 0.27$ & 0.95
           & 3,046.81 & $68.23 \pm 0.30$
           & 448.64 & $68.22 \pm 0.27$ & -0.01 \\

        15 & 317.71 & $45.26 \pm 0.52$
           & 58.55 & $45.54 \pm 0.44$ & 0.62
           & 3,047.54 & $64.87 \pm 0.63$
           & 468.72 & $64.67 \pm 0.55$ & -0.30 \\

        20 & 318.40 & $46.11 \pm 0.20$
           & 59.18 & $46.03 \pm 0.27$ & -0.17
           & 3,048.26 & $61.16 \pm 0.72$
           & 488.80 & $60.90 \pm 0.75$ & -0.43 \\
        \bottomrule
      \end{tabular}
    \end{sc}
  \end{scriptsize}
\end{table*}
Table \ref{tab:conv_cv}  demonstrates that, aside from MetaFastKANConv, the Meta variants match or exceed the classification accuracy of all the non-Meta variants at a reduced memory cost.  For example, in the case of MetaKANConv on Cifar-10, the reduced memory cost is up to $4.5\times$ (from 13.7 MB to 3.1 MB) compared to KANConv.  Upon clustering, we find that the compression factor increases by a factor of $7.1\times$ (from 3.1 MB to 435.34 KB).  In total, compared to the KANConv, the MetaClusterKANConv achieves a reduction of up to $31.7\times$ (from 13.7 MB to 435.34 KB).  For such a reduction in storage cost, after fine-tuning, we find there is a negligible impact on the downstream classification accuracy when comparing KANConv to MetaKANConv. Although the non-Meta variants recover accuracy from the fine-tuning step, the recovery still leaves them far behind their original state.  For instance, upon clustering, FastKANConv experiences up to a $2.93\times$ decrease (from 76.30\% to 26.01 \% ) in classification accuracy.  Then, when recovered with fine-tuning, it still retains a $1.32\times$ decrease (from 76.30\% to 57.45 \%) in classification accuracy.  These results again demonstrate the importance of MetaCluster for maintaining accuracy at high compression factors.

\subsection{Ablation Study}
\label{sec:ablation}

\subsubsection{Basis Coefficient Count}

We investigate the impact of the basis coefficient count (i.e., the number of radial basis functions) on downstream clustering performance.  Since our technique employs a meta-learner to learn a lower-dimensional subspace, our approach should be resilient to changes in the basis coefficients we cluster. 
From Table \ref{tab:grid_size}, we can see that the initial accuracy of MetaFastKAN and MetaFastKANConv accuracy peaks at a coefficient count of 5 before dropping.  However, even with these variations in accuracy for the coefficient count, the relative percentage change remains the same.  This indicates that, regardless of the number of coefficients, our MetaCluster 
facilitates easier clustering of weights.

\subsubsection{Cluster Count}
We verify that the MetaCluster framework scales well with the cluster count by comparing its cluster scaling properties with those of a non-meta KAN.  As shown in Figure \ref{fig:clustercount}, MetaCluster maintains accuracy well as clusters are reduced, especially after fine-tuning. For the convolutional model, decreasing the cluster count from 256 to 32 lowers MetaClusterFastKANConv accuracy by only $2\%$, whereas ClusterFastKANConv drops by $21.39\%$ over the same range.  Such results demonstrate our approach can achieve even greater compression, albeit with a minor loss in accuracy.
\begin{figure*}[htb]
  \centering
  \begin{subfigure}[b]{0.48\textwidth}
    \centering
    \begin{tikzpicture}
      \begin{axis}[
        width=\textwidth,
        height=3.75cm,
        xlabel={Number of Clusters },
        ylabel={Accuracy $(\%)$},
        grid=major,
        label style={font=\small},
        tick label style={font=\scriptsize},
        xmin=3, xmax=80,
        ymin=10, ymax=55,
        xmode=log,
        log basis x=2,
        xtick={4,8,16,32,64},
        xticklabels={$2^2$,$2^3$,$2^4$,$2^5$,$2^6$},
       legend style={
            at={(0.5,1.05)},         
            anchor=south,            
            font=\tiny,              
            row sep=-2pt,            
            column sep=5pt,          
            /tikz/every even column/.append style={column sep=5pt}
          },
          legend columns=2,          
          every axis legend/.append style={
            cells={anchor=west},     
          },
          legend image post style={scale=0.5}
      ]

        \addplot[
          red,
          thick,
          dashed,
        ] coordinates {
          (3,49.28) (80,49.28) 
        };
        \addlegendentry{FastKAN}
        
        \addplot[
          blue,
          thick,
          dashed,
        ] coordinates {
          (3,47.41)  (80,47.41) 
        };
        \addlegendentry{MetaFastKAN}

        \addplot[
          red,
          thick,
          mark=square*,
          mark options={fill=white}
        ] coordinates {
          (4,18.00) (8,17.99) (16,23.53) (32,24.38) (64,24.96) 
        };
        \addlegendentry{ClusterFastKAN}
        
        \addplot[
          blue,
          thick,
          mark=square*,
          mark options={fill=white}
        ] coordinates {
          (4,24.30) (8,44.54) (16,47.46) (32,47.65) (64,47.49) 
        };
        \addlegendentry{MetaClusterFastKAN}

        \addplot[
          red,
          thick,
          mark=triangle*,
          mark options={fill=white}
        ] coordinates {
          (4,31.81) (8,35.88) (16,38.26) (32,41.34) (64,42.32) 
        };
        \addlegendentry{\shortstack{\hspace{1em}+ fine-tune}}
        
        \addplot[
          blue,
          thick,
          mark=triangle*,
          mark options={fill=white}
        ] coordinates {
          (4,38.88) (8,46.62) (16,47.35) (32,47.52) (64,47.68) 
        };
        \addlegendentry{\shortstack{\hspace{1em}+ fine-tune}}

      \end{axis}
    \end{tikzpicture}
    \caption{Fully-connected KAN variants.}
    \label{fig:subA}
  \end{subfigure}
  \hfill
   \begin{subfigure}[b]{0.48\textwidth}
    \centering
    \begin{tikzpicture}
      \begin{axis}[
        width=\textwidth,
        xlabel={Number of Clusters },
        ylabel={Accuracy $(\%)$},
        grid=major,
        label style={font=\small},
        tick label style={font=\scriptsize},
        height=3.75cm,
        xmin=14, xmax=300,
        ymin=0, ymax=80,
        xmode=log,
        log basis x=2,
        xtick={16,32,64,128,256},
        xticklabels={$2^4$,$2^5$,$2^6$,$2^7$,$2^8$},
          legend style={
            at={(0.5,1.05)},         
            anchor=south,            
            font=\tiny,              
            row sep=-2pt,            
            column sep=5pt,          
            /tikz/every even column/.append style={column sep=5pt}
          },
          legend columns=2,          
          every axis legend/.append style={
            cells={anchor=west},     
          },
          legend image post style={scale=0.5}
      ]
        \addplot[
          red,
          thick,
          dashed,
        ] coordinates {
          (14,76.28) (300,76.28) 
        };
        \addlegendentry{FastKANConv}

        \addplot[
          blue,
          thick,
          dashed,
        ] coordinates {
          (14,71.01)  (300,71.01) 
        };
        \addlegendentry{MetaFastKANConv}
        
        \addplot[
          red,
          thick,
          mark=square*,
          mark options={fill=white}
        ] coordinates {
          (16,13.84) (32,11.77) (64,11.11) (128,12.89) (256,11.58) 
        };
        \addlegendentry{ClusterFastKANConv}

        \addplot[
          blue,
          thick,
          mark=square*,
          mark options={fill=white}
        ] coordinates {
          (16,44.88) (32,61.80) (64,67.70) (128,69.52) (256,70.19) 
        };
        \addlegendentry{MetaClusterFastKANConv}
        
        \addplot[
          red,
          thick,
          mark=triangle*,
          mark options={fill=white}
        ] coordinates {
          (16,30.95) (32,34.38) (64,44.51) (128,49.76) (256,55.77) 
        };
        \addlegendentry{\shortstack{\hspace{1em}+ fine-tune}}

        \addplot[
          blue,
          thick,
          mark=triangle*,
          mark options={fill=white}
        ] coordinates {
          (16,64.15) (32,68.82) (64,69.90) (128,70.44) (256,70.62) 
        };
        \addlegendentry{\shortstack{\hspace{1em}+ fine-tune}}

      \end{axis}
    \end{tikzpicture}
    \caption{Convolutional KAN variants.}
    \label{fig:clustercount}
  \end{subfigure}

  \caption{Classification accuracy vs. number of clusters for FastKAN model variants.
  }
  \label{fig:twographs_with_legends}
\end{figure*}

\subsubsection{Meta-learner Embedding Size}
The embedding size of the meta-learner directly influences the ease of clustering the meta-learner generated weights.  The general trend is that as the embedding dimension increases, it becomes increasingly difficult to cluster the generated weights appropriately.  While we have qualitatively demonstrated this in Figure \ref{fig:tsne}, where increasing the embedding dimension makes the plot less structured, we provide a quantitative analysis of this effect in Appendix \ref{Appendix:Embedding}.

\subsection{Computational Cost Results}
\begin{table*}[b]
	\setlength{\tabcolsep}{1.5pt} 
	\caption{Training/fine-tuning and inference times (in seconds)}
	\label{tab:train_inference_time}
	\centering
	\begin{scriptsize}
		\begin{sc}
			\begin{tabular}{lccc|ccc|ccc}
				\toprule
				\textbf{Model} & \multicolumn{2}{c}{\textbf{MNIST}} & & \multicolumn{2}{c}{\textbf{CIFAR-10}} & & \multicolumn{2}{c}{\textbf{CIFAR-100}} & \\
				\cmidrule(lr){2-3} \cmidrule(lr){5-6} \cmidrule(lr){8-9}
				& Train & Inference & & Train & Inference & & Train & Inference \\
				\midrule
                KAN & $2.155 \pm 0.047$  
                & $0.320 \pm 0.005$  
                &&
                $2.661 \pm 0.049$  
                & $0.510 \pm 0.008$ 
                &&
                $2.730 \pm 0.060$  
                & $0.509 \pm 0.007$ \\
                MetaKAN & $2.501 \pm 0.055$  
                & $0.337 \pm 0.003$  
                &&
                $2.963 \pm 0.056$  
                & $0.538 \pm 0.012$ 
                &&
                $2.956 \pm 0.096$  
                & $0.526 \pm 0.008$ \\
                MetaClusterKAN & $2.284 \pm 0.048$ 
                & $0.323 \pm 0.002$  
                &&
                $3.259 \pm 0.045$  
                & $0.521 \pm 0.008$ 
                &&
                $3.252 \pm 0.091$  
                & $0.542 \pm 0.019$ \\
                \midrule
                
                KANConv & $9.158 \pm 0.184$ 
                & $0.922 \pm 0.009$  
                &&
                $10.105 \pm 0.085$  
                & $1.130 \pm 0.003$ 
                &&
                $10.370 \pm 0.162$  
                & $1.204 \pm 0.023$ \\
                MetaKANConv & $10.039 \pm 0.228$ 
                & $0.930 \pm 0.009$   
                &&
                $11.024 \pm 0.080$   
                & $1.184 \pm 0.025$  
                &&
                $10.946 \pm 0.097$   
                & $1.223 \pm 0.011$ \\
                
                MetaClusterKANConv  & $9.931 \pm 0.160$  
                & $0.951 \pm 0.017$  
                &&
                $10.605 \pm 0.144$  
                & $1.149 \pm 0.032$ 
                &&
                $10.929 \pm 0.156$  
                & $1.202 \pm 0.004$ \\
				\bottomrule
			\end{tabular}
		\end{sc}
	\end{scriptsize}
\end{table*}
\begin{wraptable}{R}{0.5\columnwidth}
    \setlength{\tabcolsep}{1.5pt} 
    \caption{Clustering times (in seconds)}
    \label{tab:cluster_time}
    \centering
    \begin{scriptsize}
        \begin{sc}
            \begin{tabular}{lccc}
                \toprule
                \textbf{Model} & \textbf{MNIST} & \textbf{CIFAR-10} & \textbf{CIFAR-100} \\
                \midrule
                KAN      & $0.029 \pm 0.011$ & $0.059 \pm 0.007$ & $0.080 \pm 0.022$ \\
                MetaKAN  & $0.024 \pm 0.012$ & $0.048 \pm 0.011$ & $0.062 \pm 0.011$ \\ 
                \midrule
                KANConv & $3.587 \pm 0.145$ & $3.650 \pm 0.098$ & $3.542 \pm 0.114$ \\
                MetaKANConv & $2.806 \pm 0.072$ & $2.812 \pm 0.067$ & $2.791 \pm 0.092$ \\
                \bottomrule
            \end{tabular}
        \end{sc}
    \end{scriptsize}
\end{wraptable}
\label{sec: computation}
As our approach is a three-stage framework consisting of a training stage, a clustering stage, and a fine-tuning stage, each stage will contribute to the computational cost during training.  Furthermore, because our approach stores the KAN architecture as a codebook and indices that index the codebook, the forward inference cost also incurs computational overhead.  We quantify the clustering stage cost in Table \ref{tab:cluster_time} and training/fine-tuning and inference costs in Table \ref{tab:train_inference_time}.  Each metric is reported in seconds, averaged over 10 runs, with standard error, on an RTX 5090 GPU.  The training and inference times are reported as the time to complete one epoch and the time to evaluate on the test set, respectively.  

From observing Table \ref{tab:cluster_time} we can see that compared to training times the clustering times are negligible (e.g., 0.029 s once vs 2.155 s per epoch).  We can also see that the time required to train MetaKAN per epoch is slightly longer than KAN.  Similarly, during fine-tuning (which runs for much fewer epochs than the training stage), the time per epoch is slightly longer for MetaClusterKAN (an identical architecture to ClusterKAN) than for KAN.  Such a trend deviates in inference, where the MetaClusterKAN is as fast as KAN. Similar trends are seen with the convolutional models. Such results show there is minimal overhead from MetaCluster. 

\subsection{Equation Modeling Results}
\label{sec: equation_modeling}
Our high-dimensional equation modeling results examined MetaCluster's ability to model 1000-dimensional equations bounded on $[-1,1]$. All KAN models used were identical to our fully-connected image classification experiments.  We explored the equations $f_1(x) = \exp\left(\frac{1}{n} \sum \sin^2\left(\frac{\pi x}{2}\right)\right)$, $f_2(x) = \sum x^2 + x^3$, $f_3(x) = \exp\left(-\frac{1}{n}\sum x^2\right)$ following a similar procedure to \citet{zhao2025improving}. The clustered models used a cluster size of 4.  Each reported MSE result is an average over 10 runs. We provide the complete set of hyperparameters in Appendix \ref{app:equation_hyper}.  
\begin{wraptable}{r}{0.6\columnwidth}
    \setlength{\tabcolsep}{2pt}
    \vspace{0.7cm}
    \caption{High-dimensional equation modeling MSE and memory (KB) footprint comparison.}
    \label{tab:equation}
    \centering
    \begin{scriptsize}
    \begin{sc}
    \begin{tabular}{l c | c | c | c}
        \toprule
        \textbf{Model} & \textbf{Memory} 
        & \textbf{$f_1(x)$ MSE} 
        & \textbf{$f_2(x)$ MSE} 
        & \textbf{$f_3(x)$ MSE} \\
        \midrule
        KAN & 1303.6 
            & \(5.06\times10^{-5}\)
            & \(2.70\times10^{-1}\)
            & \(5.30\times10^{-7}\) \\
        
        MetaKAN & 181.6
            & \(3.05\times10^{-5}\)
            & \(2.41\times10^{0}\)
            & \(2.76\times10^{-6}\) \\
        
        ClusterKAN & 10.5
            & \(9.85\times10^{-1}\)
            & \(1.66\times10^{2}\)
            & \(1.27\times10^{-1}\) \\
        
        \hspace{2em}+ fine-tune & 10.5
            & \(2.24\times10^{-4}\)
            & \(8.92\times10^{-1}\)
            & \(7.82\times10^{-6}\) \\
        
        MetaClusterKAN & 10.5
            & \(9.03\times10^{-4}\)
            & \(1.58\times10^{0}\)
            & \(3.75\times10^{-4}\) \\
        
        \hspace{2em}+ fine-tune & 10.5
            & \(2.70\times10^{-5}\)
            & \(5.04\times10^{-1}\)
            & \(7.55\times10^{-7}\) \\
        
        \bottomrule
    \end{tabular}
    \end{sc}
    \end{scriptsize}
\end{wraptable}

From Table \ref{tab:equation}, we can see that MetaClusterKAN achieves identical or, in some cases, reduced mean squared error (MSE) than the original KAN model after clustering. MetaClusterKAN can achieve these equation modeling capabilities with a 124.1x reduction in memory.  This is not the case for ClusterKAN, which shows a significant increase in MSE after clustering (from $5.061\times10^{-5}$ to $2.239\times10^{-4}$).  These results demonstrate that the principles from MetaCluster extend beyond image classification into the equation modeling domain. Furthermore, as detailed in Appendix \ref{app:pde}, this structural resilience also translates to physics-informed neural networks, where MetaCluster successfully preserves the precision required to solve partial differential equations despite memory compression.  

\section{Related Works}
\subsection{Kolmogorov-Arnold Networks}
KANs have evolved rapidly, with numerous variants exploring different basis functions such as B-splines, radial basis functions (RBFs), Chebyshev and Legendre polynomials, Gram polynomials, wavelets, and rational functions \citep{liu2024kan, li2024kolmogorov, ss2024chebyshev, bozorgasl2024wavkan, aghaei2024rkan}. These architectures have demonstrated strong performance in scientific computing and equation modeling \citep{li2025u, wang2025kolmogorov, coffman2025matrixkan, koenig2024kan} and have recently extended to computer vision through designs such as the Kolmogorov-Arnold Transformer \citep{yang2024kolmogorov, raffel2025flashkat}. Despite these successes, KANs face persistent challenges, including training instability, computational overhead, and a substantial increase in parameter count compared to MLPs \citep{yu2024kan,VisionKAN2024}.  Our work directly addresses the memory-scaling bottleneck of KANs by creating topologically identical KANs that are smaller than comparable MLPs while maintaining accuracy.

\subsection{HyperNetworks}
Hypernetworks reduce trainable parameter counts by replacing task or instance-specific weights with a shared generator that predicts them. The idea originates in early meta-learning work such as \citet{ba2016using}, which uses a small network to generate classifier weights from context, and \citet{bertinetto2016learning}, which learns to emit a one-shot tracker’s parameters conditioned on a single exemplar. \citet{ha2016hypernetworks} then formalized Hypernetworks for CNNs and RNNs, showing a compact hypernetwork can match the accuracy of a standard model while drastically cutting the number of directly optimized parameters. Later extensions use task embeddings to condition a single hypernetwork for multi-task or multi-objective weight generation \citep{savarese2019learning, navon2020learning} . 

MetaKANs \citep{zhao2025improving} extend this paradigm to KANs by observing that the dominant cost in a KAN is the storage of the coefficients for each univariate activation. Instead of optimizing all coefficients directly, a small meta-learner maps per-activation embeddings to basis coefficients, capturing a shared rule for weight generation across activations \citep{zhao2025improving}. In contrast, our method uses the meta-learner only during training to impose a clusterable geometry on per-edge coefficient embeddings.  Then, at inference, we dispense with any hypernetwork, incurring zero runtime overhead while achieving even greater parameter efficiency.

\subsection{Weight Sharing}
Weight sharing compresses networks by restricting each layer’s parameters to a small set of shared centroids and storing a codebook, along with per-weight indices. Scalar quantization approaches, popularized by \citet{han2015deep} introduced a pipeline combining of magnitude pruning and K-means weight sharing. Subsequent theory linked quantization error to loss curvature and leveraged a  Hessian-weighted K-means to cluster \citep{choi2016towards}. 
Evolving from scalar quantization, vector quantization offers a natural extension by clustering weight vectors rather than individual scalars, which early work applied to improve compression–accuracy trade-offs \citep{gong2014compressing}. More recently, the rise of large language models has driven specialized vector quantization approaches, which leverage Hessian information and the inherent sparsity of LLMs to optimize vector-level assignments \citep{liu2024vptq, egiazarian2024extreme,van2024gptvq, tseng2024quip}. We demonstrate in Appendix \ref{app:gptvq} that such Hessian information is insufficient for KANs, a result that is in line with Hessian-aware quantization techniques reported in \citet{fuad2025quantkan}. Building on vector quantization research, our MetaCluster framework is the first to successfully apply weight-sharing principles to the KAN. Crucially, our approach deviates from standard vector quantization paradigms: whereas traditional methods typically cluster across the external degrees of freedom between neurons or layers, MetaCluster targets the internal degrees of freedom within the per-edge activation functions themselves.

It is worth noting that, while weight-sharing is one avenue for model compression, bit-width quantization \citep{fuad2025quantkan} is an alternative avenue, which is largely orthogonal and complementary to the proposed approach.  Applying quantization on top of MetaCluster (e.g., quantizing the codebook of centroids) can further reduce the memory footprint.   

\section{Conclusion}
\label{sec:conclusion}
We introduced MetaCluster, a compression framework that makes KANs scalable by combining meta-learned manifold shaping with weight sharing. A lightweight meta-learner maps low-dimensional embeddings to per-edge basis coefficients, thereby constraining KAN activations to lie on a compact manifold amenable to clustering. We then apply K-means in coefficient space, replace per-edge parameters with codebook centroids and compact indices, discard the meta-learner, and briefly fine-tune centroids to recover any loss. This design directly targets the dimensionality driver of KAN memory, yielding a storage advantage that grows with the number of coefficients per edge. 

Evaluation shows that MetaCluster achieves up to $80\times$ and $124.1\times$ reductions in parameter storage for image classification and equation modeling, respectively, with competitive accuracy compared to uncompressed baselines. While this work validates the framework on standard benchmarks, establishing formal theoretical guarantees and extending these storage gains to large-scale architectures remain promising directions for future research. Ultimately, MetaCluster facilitates the deployment of KANs in resource-constrained environments, offering a path toward more accessible AI without introducing unique ethical risks.

\bibliographystyle{plainnat}
\bibliography{example_paper}

\newpage
\appendix
\onecolumn
\section{Appendix}

\subsection{Hyperparameters}
\label{Appendix: Hyperparameters}
\subsubsection{Image Classification Fully-Connected Setup}
\label{sec:fc_architecture}
For all image classification experiments, we did a 90/10 training/validation split of the training data.  Our fully-connected architecture follows \citet{zhao2025improving} in that we stack two fully-connected KAN layers, with a 32-dimensional hidden state between them.  Each KAN variant uses a SiLU activation \citep{elfwing2018sigmoid}.  The meta-learner variants contained a meta-learner with a hidden dimension of 32. Concretely:
\begin{itemize}[noitemsep, topsep=0pt]
  \item \textbf{KAN:} degree–3 B-splines, grid range $[-1,1]$, grid size $5$ \citep{liu2024kan}.
  \item \textbf{FastKAN:} $8$ RBFs over $[-2,2]$ \citep{li2024kolmogorov}.
  \item \textbf{GramKAN:} degree–3 polynomial \citep{drokin2024kolmogorov}.
\end{itemize}

We train with AdamW 
\citep{loshchilov2017decoupled} and for at most 50 epochs, with early stopping (patience = 10).  After the final epoch, we cluster the learned weights into 16 groups, then fine-tune for an additional 5 epochs.
For our fully-connected architecture experiments Table \ref{tab:hyper_kan} reports the full set of hyperparameters for our KAN variants and Table \ref{tab:hyper_meta_kan} reports the  full set of hyperparameters for our MetaKAN variants.

\begin{table}[ht]
\centering
\caption{Hyperparameters for fully-connected KAN variants.}
\label{tab:hyper_kan}
\begin{scriptsize}
\begin{sc}
\begin{tabular}{lccc}
\toprule
Hyperparameter & KAN & FastKAN & GramKAN \\
\midrule
Hidden dimension & 32 & 32 & 32 \\
Activation & SiLU & SiLU & SiLU \\
Degree & 3 (Spline) & - & 3 \\
Grid range & $[-1,1]$ & $[-2,2]$ & - \\
Grid size & 5 & 8(Num grids) & - \\
Optimizer & AdamW & AdamW & AdamW \\
Learning rate & $1\times10^{-4}$ & $1\times10^{-3}$ & $1\times10^{-3}$ \\
\shortstack{Learning rate for \\finetuning (lr\_c)} & $1\times10^{-4}$ & $1\times10^{-4}$ & $1\times10^{-4}$ \\
Batch size & 128 & 128 & 128 \\
Epochs & 50 & 50 & 50 \\
Early stopping patience & 10 & 10 & 10 \\
Clustered training epochs & 5 & 5 & 5 \\
Early stopping patience (fine-tuning) & 3 & 3 & 3 \\
Number of clusters & 16 & 16 & 16 \\
\bottomrule
\end{tabular}
\end{sc}
\end{scriptsize}
\end{table}

\begin{table}[ht]
\centering
\caption{Hyperparameters for fully-connected MetaKAN variants.}
\label{tab:hyper_meta_kan}
\begin{scriptsize}
\begin{sc}
\begin{tabular}{lccc}
\toprule
Hyperparameter & MetaKAN & MetaFastKAN & MetaGramKAN \\
\midrule
Hidden dimension & 32 & 32 & 32 \\
Embedding dimension & 1 & 1 & 1 \\
Activation & SiLU & SiLU & SiLU \\
Degree & 3 (Spline) & - & 3 \\
Grid range & $[-1,1]$ & $[-2,2]$ & - \\
Grid size & 5 & 8(Num-grids) & - \\
Optimizer set & double & double & double \\
Optimizer & AdamW & AdamW & AdamW \\
Learning rate for meta-learner (lr\_h) & $5\times10^{-4}$ & $1\times10^{-3}$ & $1\times10^{-4}$ \\
Learning rate for embeddings (lr\_c)& $5\times10^{-3}$ & $1\times10^{-2}$ & $1\times10^{-3}$ \\
\shortstack{Learning rate for \\finetuning (lr\_c)} & $1\times10^{-4}$ & $1\times10^{-4}$ & $1\times10^{-4}$ \\
Batch size & 128 & 128 & 128 \\
Epochs & 50 & 50 & 50 \\
Clustered training epochs & 5 & 5 & 5 \\
Early stopping patience & 10 & 10 & 10 \\
Early stopping patience (fine-tuning) & 3 & 3 & 3 \\
Number of clusters & 16 & 16 & 16 \\
\bottomrule
\end{tabular}
\end{sc}
\end{scriptsize}
\end{table}

\subsubsection{Image Classification Convolutional Setup}
\label{sec:conv_architecture}
For all image classification experiments, we did a 90/10 training/validation split of the training data. We apply data augmentation to the train set for the convolutional architecture, following an identical procedure to \citet{drokin2024kolmogorov}. Our convolutional architecture is taken from \citet{drokin2024kolmogorov}: four convolutional layers with channel progression $[32,64,128,512]$, each using $3\times 3$ kernels, stride 1, and padding 1.  As with the fully-connected experiments, each KAN variant uses a SiLU activation \citep{elfwing2018sigmoid} and the meta-learner versions use a meta-learner with a hidden dimension of 32. We replace the usual activation with our kernels:

\begin{itemize}[noitemsep, topsep=0pt]
  \item \textbf{KANConv:} degree–3 B-splines, grid range $[-3,3]$, grid size $5$ \citep{liu2024kan}.
  \item \textbf{FastKANConv:} $8$ RBFs on $[-3,3]$ \citep{li2024kolmogorov}.
  \item \textbf{GramKANConv:} degree–3 polynomial \citep{drokin2024kolmogorov}.
\end{itemize}

Training again uses AdamW 
for up to 150 epochs with early stopping (patience = 10).  Final weights are clustered into 16 groups and fine-tuned for 5 epochs. 
For our convolutional architecture experiments Table \ref{tab:hyper_kan_conv} reports the full set of hyperparameters for our KAN variants and Table \ref{tab:hyper_meta_kan_conv} reports the full set of hyperparameters for our MetaKAN variants.

\begin{table}[ht]
\centering
\caption{Hyperparameters for convolutional non-meta KAN variants.}
\label{tab:hyper_kan_conv}
\begin{scriptsize}
\begin{sc}
\begin{tabular}{lccc}
\toprule
Hyperparameter & KAN & FastKAN & KAGN \\
\midrule
Hidden dimension & [32, 64, 128, 256] & [32, 64, 128, 256] & [32, 64, 128, 256] \\
Activation & SiLU & SiLU & SiLU \\
Degree & 3 (Spline) & - & 3 \\
Grid range & $[-3,3]$ & $[-3,3]$ & - \\
Grid size & 5 & 8(Num-Grids) & - \\
Dropout & 0.25 & 0.25 & 0.25 \\
Dropout (linear layers) & 0.5 & 0.5 & 0.5 \\
Optimizer & AdamW & AdamW & AdamW \\
\shortstack{Learning rate for \\finetuning (lr\_c)} & $1\times10^{-5}$ & $1\times10^{-5}$ & $1\times10^{-5}$ \\
learning rate (lr) & $1\times10^{-3}$ & $1\times10^{-3}$ & $1\times10^{-3}$ \\
Batch size & 128 & 128 & 128 \\
Epochs & 150 & 150 & 150 \\
Clustered training epochs & 5 & 5 & 5 \\
Early stopping patience & 10 & 10 & 10 \\
\shortstack{Early stopping patience\\ (fine-tuning)} & 3 & 3 & 3 \\
Number of clusters & 256 & 256 & 256 \\
Convolution Groups & 1 & 1 & 1 \\

\bottomrule
\end{tabular}
\end{sc}
\end{scriptsize}
\end{table}

\begin{table}[ht]
\centering
\caption{Hyperparameters for convolutional MetaKAN variants.}
\label{tab:hyper_meta_kan_conv}
\begin{scriptsize}
\begin{sc}
\begin{tabular}{lccc}
\toprule
Hyperparameter & MetaKAN & MetaFastKAN & MetaKAGN \\
\midrule
Hidden dimension & [32, 64, 128, 256] & [32, 64, 128, 256] & [32, 64, 128, 256] \\
Embedding dimension & 2 & 2 & 1 \\
Activation & SiLU & SiLU & SiLU \\
Degree & 3 (Spline) & - & 3 \\
Grid range & $[-3,3]$ & $[-3,3]$ & - \\
Grid size & 5 & 8(Num-grid) & - \\
Optimizer set & double & double & double \\
Optimizer & AdamW & AdamW & AdamW \\
\shortstack{Learning rate \\for meta-learner (lr\_h)} & $1\times10^{-4}$ & $1\times10^{-4}$ & $1\times10^{-4}$ \\
Learning rate embedding (lr\_e) & $5\times10^{-3}$ & $5\times10^{-3}$ & $5\times10^{-3}$ \\
\shortstack{Learning rate for \\finetuning (lr\_c)} & $1\times10^{-5}$ & $1\times10^{-5}$ & $1\times10^{-5}$ \\
Global learning rate (lr) & $1\times10^{-3}$ & $5\times10^{-3}$ & $5\times10^{-3}$ \\
Batch size & 128 & 128 & 128 \\
Epochs & 150 & 150 & 150 \\
Clustered training epochs & 5 & 5 & 5 \\
Early stopping patience & 10 & 10 & 10 \\
\shortstack{Early stopping patience \\(fine-tuning)} & 3 & 3 & 3 \\
Number of clusters & 256 & 256 & 256 \\
Embedding scheduler & Yes & Yes & Yes \\
Hypernet scheduler & Yes & Yes & Yes \\
Dropout & 0.25 & 0.25 & 0.25 \\
Dropout (linear layers) & 0.5 & 0.5 & 0.5 \\
Convolution Groups & 1 & 1 & 1 \\
\bottomrule
\end{tabular}
\end{sc}
\end{scriptsize}
\end{table}

\subsubsection{Equation Modeling Setup}
\label{app:equation_hyper}
We report our complete set of hyperparameter for our equation modeling experiments in Table \ref{tab:equation_modeling_hyper}.
\begin{table}[ht]
\centering
\caption{Hyperparameters for fully-connected KAN equation modeling experiments.}
\label{tab:equation_modeling_hyper}
\begin{scriptsize}
\begin{sc}
\begin{tabular}{lc}
\toprule
Hyperparameter & KAN  \\
\midrule
Hidden dimension & 32  \\
Activation & SiLU \\
Degree & 3 (Spline)  \\
Grid range & $[-1,1]$  \\
Grid size & 5  \\
Optimizer & AdamW  \\
Learning rate & $1\times10^{-4}$ \\
\shortstack{Learning rate for \\finetuning (lr\_c)} & $1\times10^{-4}$ \\
Batch size & 256 \\
Epochs & 100 \\
Clustered training epochs & 10 \\
Number of clusters & 4 \\
Domain of datapoints & [-1,1] \\
Train Set samples & 50,000 \\
Test Set samples & 20,000 \\
\bottomrule
\end{tabular}
\end{sc}
\end{scriptsize}
\end{table}

\subsection{MetaCluster vs. GPTVQ}
\label{app:gptvq}
To further validate our approach, we benchmark MetaCluster against GPTVQ \citep{van2024gptvq}, a state-of-the-art Hessian-aware vector quantization technique, across fully-connected KAN, FastKAN, and GramKAN architectures. As detailed in Table \ref{tab:gptvq}, MetaCluster consistently demonstrates a superior accuracy-compression trade-off at identical memory footprints. Most notably, MetaCluster preserves structural integrity immediately after clustering, whereas GPTVQ suffers from catastrophic performance degradation. For example, MetaClusterFastKAN on MNIST retains 94.87\% accuracy before fine-tuning, whereas GPTVQ collapses to 30.80\%. Although a brief fine-tuning stage helps GPTVQ recover a significant portion of its lost accuracy, it ultimately fails to match MetaClusterFastKAN's final performance, particularly on more complex datasets like CIFAR-100, where the fine-tuned MetaClusterFastKAN achieves 19.05\% accuracy compared to GPTVQ's 11.68\%. These empirical results indicate that Hessian-aware clustering is suboptimal for the high-dimensional activation coefficient space of KANs, and that explicitly constraining weights to a low-dimensional manifold via MetaCluster provides a fundamentally better initialization for weight sharing, avoiding severe accuracy loss.
\begin{table*}[h]
	\setlength{\tabcolsep}{3pt} 
	\caption{Classification accuracy and memory (KB) comparison of a fully-connected network between GPTVQ and MetaCluster.}
	\label{tab:gptvq}
	\centering
	\begin{scriptsize}
		\begin{sc}
			\begin{tabular}{lccc|ccc|ccc}
				\toprule
				\textbf{Model} & \multicolumn{2}{c}{\textbf{MNIST}} & & \multicolumn{2}{c}{\textbf{CIFAR-10}} & & \multicolumn{2}{c}{\textbf{CIFAR-100}} & \\
				\cmidrule(lr){2-3} \cmidrule(lr){5-6} \cmidrule(lr){8-9}
				& Memory & Acc. & & Memory & Acc. & & Memory & Acc. \\
				\midrule
				MetaClusterKAN & 13.84 & $96.06 \pm 0.10 $ & & 38.34 & $45.47 \pm 0.44$ & & 39.75 & $18.97 \pm 0.15$ \\
                \hspace{2em}+ fine-tune & 13.84 & $96.41 \pm 0.10$ & & 38.34 & $45.95 \pm 0.33$ & & 39.75 & $19.16 \pm 0.17$ \\
                GPTVQ & 13.84 & $80.96 \pm 2.60$ & & 38.34 & $31.38 \pm 3.73$ & & 39.75 & $3.49 \pm 0.28$ \\
                \hspace{2em}+ fine-tune & 13.84 & $92.61 \pm 0.28$ &  & 38.34 & $43.47 \pm 0.13$ & & 39.75 & $11.21 \pm 0.50$\\
                \midrule
				MetaClusterFastKAN & 20.54 & $94.87 \pm 0.29$ & & 57.30 & $46.96 \pm 0.44$ & & 58.71 & $18.66 \pm 0.48$ \\
                \hspace{2em}+ fine-tune & 20.54 & $95.59 \pm 0.18$ & & 57.30 & $47.60 \pm 0.37$ & & 58.71 & $19.05 \pm 0.47$ \\
                GPTVQ & 20.54 & $30.80 \pm 2.24$ & & 57.30 & $24.57 \pm 0.93$ & & 58.71 & $7.13 \pm 0.31$ \\
                \hspace{2em}+ fine-tune & 20.54 & $87.16 \pm 0.61$ &  & 57.30 & $39.10 \pm 0.0.32$ &  & 58.71 & $11.68 \pm 0.32$\\
                \midrule
				MetaClusterGramKAN & 14.15 & $93.80 \pm 0.72$ & & 38.65 & $47.66 \pm 0.25$ & & 40.76 & $18.18 \pm 0.20$\\
                \hspace{2em}+ fine-tune & 14.15 & $95.32 \pm 0.12$ & & 38.65 & $48.47 \pm 0.30$ & & 40.76 & $18.96 \pm 0.25$\\
                GPTVQ & 13.96 & $82.81 \pm 2.23$ & & 38.46 & $32.44 \pm 1.58$ & & 40.58 & $7.49 \pm 0.56$ \\
                \hspace{2em}+ fine-tune & 13.96 & $95.44 \pm 0.14$ &  & 38.46 & $45.48 \pm 0.57$ &  & 40.58 & $14.02 \pm 0.63$\\
				\bottomrule
			\end{tabular}
		\end{sc}
	\end{scriptsize}
\end{table*}

\subsection{Partial Differential Equations}
\label{app:pde}
To evaluate the compressibility of KAN and MetaKAN in the physics-informed setting, we train both architectures as PINNs on four canonical PDEs with known analytical solutions: the heat equation ($\partial_t u = 0.1\,\partial_{xx} u$), the wave equation ($\partial_{tt} u = \partial_{xx} u$), the advection equation ($\partial_t u + \partial_x u = 0$), and the Laplace equation ($\partial_{xx} u + \partial_{yy} u = 0$), all defined on $[0,1]^2$ with Dirichlet boundary conditions. We enforce boundary and initial conditions exactly using the hard constraint formulation \citep{lagaris1998artificial}, expressing the solution as $u = B + D \cdot N$ where $B$ satisfies the prescribed conditions, $D$ vanishes on constrained boundaries, and $N$ is the network output. Specifically, these exact constraints are formulated as $u(x,t) = \sin(\pi x) + x(1-x)t \cdot N(x,t)$ for the heat equation, $u(x,t) = \sin(\pi x)(1 - t^2/2) + x(1-x)t^2 \cdot N(x,t)$ for the wave equation, $u(x,t) = \sin(\pi(x-t)) + xt \cdot N(x,t)$ for the advection equation, and $u(x,y) = y\sin(\pi x) + x(1-x)y(1-y) \cdot N(x,y)$ for the Laplace equation. By satisfying the initial and boundary conditions by construction, this approach reduces the training objective to the PDE residual alone.

Both models use a KAN architecture with layers $[2, 64, 64, 1]$, B-spline order $k{=}3$, grid size $G{=}5$, and grid range $[-1,1]$. Training consists of 10{,}000 AdamW epochs with cosine annealing followed by 500 L-BFGS refinement steps on 3{,}000 fixed collocation points \citep{loshchilov2017decoupled, liu1989limited}. For MetaKAN, we use a dual-optimizer setup with separate learning rates for the MetaNet ($10^{-3}$) and embeddings ($10^{-3}$). After training, we compress each model by clustering the per-edge spline coefficient vectors (dimension $G{+}k{+}1{=}9$) into 16 centroids via K-means.  We evaluate the $L^2$ relative error on a $100 \times 100$ grid at three stages: before compression, after clustering (no fine-tuning), and after fine-tuning the cluster centroids for 5{,}000 epochs.

\begin{table*}[h]
	\setlength{\tabcolsep}{2pt} 
	\caption{L2 metric and memory (KB) comparison for PDE modeling (clusters=16).}
	\label{tab:pde_comparison}
	\centering
	\begin{scriptsize}
		\begin{sc}
			\begin{tabular}{lcc|cc|cc|cc}
				\toprule
				\textbf{Model} & \multicolumn{2}{c}{\textbf{Heat Equation}} & \multicolumn{2}{c}{\textbf{Wave Equation}} & \multicolumn{2}{c}{\textbf{Advection}} & \multicolumn{2}{c}{\textbf{Laplace}} \\
				\cmidrule(lr){2-3} \cmidrule(lr){4-5} \cmidrule(lr){6-7} \cmidrule(lr){8-9}
				& Memory & L2 Error & Memory & L2 Error & Memory & L2 Error & Memory & L2 Error \\
				\midrule
                KAN & 150.80 & $7.29\times10^{-6}$ & 150.80 & $2.85\times10^{-5}$ & 150.80 & $7.32\times10^{-6}$ & 150.80 & $3.00\times10^{-6}$ \\
                ClusterKAN & 3.80 & $1.61\times10^{-1}$ & 3.80 & $2.79\times10^{-1}$ & 3.80 & $4.91\times10^{-2}$ & 3.80 & $5.87\times10^{-2}$ \\
                \hspace{2em}+ fine-tune & 3.80 & $1.81\times10^{-4}$ & 3.80 & $1.04\times10^{-3}$ & 3.80 & $6.11\times10^{-5}$ & 3.80 & $1.18\times10^{-4}$ \\
                \midrule
                MetaKAN & 21.00 & $3.19\times10^{-5}$ & 21.00 & $9.03\times10^{-5}$ & 21.00 & $3.92\times10^{-6}$ & 21.00 & $8.37\times10^{-6}$ \\
                MetaClusterKAN & 3.80 & $9.51\times10^{-3}$ & 3.80 & $5.52\times10^{-3}$ & 3.80 & $5.28\times10^{-3}$ & 3.80 & $5.97\times10^{-3}$ \\
                \hspace{2em}+ fine-tune & 3.80 & $2.26\times10^{-5}$ & 3.80 & $1.05\times10^{-4}$ & 3.80 & $3.48\times10^{-6}$ & 3.80 & $1.82\times10^{-5}$ \\
				\bottomrule
			\end{tabular}
		\end{sc}
	\end{scriptsize}
\end{table*}

The results on the PDEs demonstrate that MetaCluster enables extreme memory compression without sacrificing the high-fidelity precision required for physics-informed neural networks. While ClusterKAN successfully reduces the model's memory footprint by nearly 40$\times$ (from 150.80 KB down to 3.80 KB), it causes severe and irreversible degradation to predictive performance. For example, on the Heat equation, the fine-tuned ClusterKAN only recovers an $L^2$ error of $1.81 \times 10^{-4}$, which is significantly worse than the baseline KAN's $7.29 \times 10^{-6}$. In contrast, after fine-tuning, MetaClusterKAN achieves the exact same 3.80 KB compressed footprint but successfully recovers baseline accuracy across all evaluated equations, even slightly outperforming the baseline KAN on the Advection equation ($3.48 \times 10^{-6}$ compared to $7.32 \times 10^{-6}$). This confirms that the proposed manifold-shaping approach is not just beneficial but fundamentally essential for extending effective weight sharing to scientific modeling tasks. 

\subsection{Ablations Extended}
\label{Appendix:ablation}
We present this extended ablation study to provide a more detailed analysis of the impact of various hyperparameters on accuracy and the effectiveness of fine-tuning across different settings. These extended results provide further insight into how variations in embedding size and coefficient count, along with fine-tuning, influence performance differences in clustering across the different KAN model variants and network architectures. 
\subsubsection{Basis Coefficient Count Extended}
Table \ref{tab:gridsizeextended} provides detailed insight into the influence of the fine-tuning stage of the MetaCluster framework on different basis coefficient counts (i.e. the grid size or number of radial basis functions).  We observe that because the majority of the original classification accuracy is preserved immediately after clustering, the fine-tuning stage yields only marginal additional gains.

\begin{table*}[ht]
\centering
\caption{Detailed results of grid sizes versus accuracy and memory for both fully-connected and convolutional MetaFastKAN networks on Cifar-10.}
\resizebox{\textwidth}{!}{%
\begin{tabular}{l c c c c c c c c c}
\hline
\textbf{Model} & 
\shortstack{\textbf{Grid Size}} & 
\shortstack{\textbf{Grid Range}} & 
\shortstack{\textbf{Embed}\\\textbf{Dim}} & 
\shortstack{\#\\\textbf{Clusters}} & 
\shortstack{\textbf{Model}\\\textbf{Accuracy}} & 
\shortstack{\textbf{Clustered}\\ \textbf{Accuracy}} & 
\shortstack{\textbf{Fine-tuned}\\\textbf{Accuracy}} & 
\shortstack{\textbf{Mem Before} \\ \textbf{Clustering (KB)}} & 
\shortstack{\textbf{Mem After} \\ \textbf{Clustering (KB)}} \\ 
\hline
\\
MetaClusterFastKAN & 5  & -2,2 & 1  & 16  & 47.41 & 47.46 & 47.35 & 316.35 & 57.30  \\
MetaClusterFastKAN & 8  & -2,2 & 1  & 16  & 47.62 & 46.96 & 47.80 & 316.76 & 57.68  \\
MetaClusterFastKAN & 10 & -2,2 & 1  & 16  & 46.91 & 47.04 & 47.49 & 317.03 & 57.93  \\
MetaClusterFastKAN & 15 & -2,2 & 1  & 16  & 45.88 & 45.15 & 45.74 & 317.71 & 58.55 \\
MetaClusterFastKAN & 20 & -2,2 & 1  & 16  & 43.57 & 43.75 & 44.31 & 318.40 & 59.18  \\
\hline
\\
MetaClusterFastKANConv & 5  & -3,3 & 2 & 256 & 71.42 & 71.06 & 71.38 & 3,046.08 & 428.55  \\
MetaClusterFastKANConv & 8  & -3,3 & 2 & 256 & 69.61 & 68.66 & 69.24 & 3,046.52 & 440.61  \\
MetaClusterFastKANConv & 10 & -3,3 & 2 & 256 & 66.78 & 66.58 & 66.74 & 3,046.81 & 448.64  \\
MetaClusterFastKANConv & 15 & -3,3 & 2 & 256 & 64.24 & 63.99 & 64.33 & 3,047.54 & 468.72  \\
MetaClusterFastKANConv & 20 & -3,3 & 2 & 256 & 54.41 & 54.45 & 54.48 & 3,048.26 & 488.80  \\
\hline
\end{tabular}}
\label{tab:gridsizeextended}
\end{table*}

\subsubsection{Meta-learner Embedding Size Extended}
\label{Appendix:Embedding}
Our ablation exploring the influence of the embedding dimension aims to quantify the influence of the embedding size on the downstream clustering performance.  We report these results in Table \ref{tab:embedsizeextended}.  From Table \ref{tab:embedsizeextended}, we can see that in both the fully-connected and convolutional architectures, as the embedding dimension increases, the classification accuracy after clustering decreases.  This validates our assumption, which we developed based on Figure \ref{fig:tsne}, that finding a lower-dimensional subspace improves downstream clustering performance. During the fine-tuning stage, we can recover most of the accuracy decrease experienced by choosing a higher-dimensional embedding.  Furthermore, the higher-dimensional choice of embedding does not affect the MetaCluster model memory footprint.  
\begin{table*}[ht]
\centering
\caption{Detailed results of embedding dimension versus accuracy for both fully-connected and convolutional MetaFastKAN networks on Cifar-10.}
\resizebox{\textwidth}{!}{%
\begin{tabular}{l c c c c c c c c c}
\hline
\textbf{Model} & 
\shortstack{\textbf{Num-grid}} & 
\shortstack{\textbf{Grid Range}} & 
\shortstack{\textbf{Embed}\\\textbf{Dim}} & 
\shortstack{\#\\\textbf{Clusters}} & 
\shortstack{\textbf{Model}\\\textbf{Accuracy}} & 
\shortstack{\textbf{Clustered}\\ \textbf{Accuracy}} & 
\shortstack{\textbf{Fine-tuned}\\\textbf{Accuracy}} & 
\shortstack{\textbf{Mem Before} \\ \textbf{Clustering (KB)}} & 
\shortstack{\textbf{Mem After} \\ \textbf{Clustering (KB)}} \\
\hline
\\
MetaClusterFastKAN & 5 & -2,2 & 4 & 16 & 48.66 & 38.28 & 44.63 & 1,202.50 & 57.30  \\
MetaClusterFastKAN & 5 & -2,2 & 3 & 16 & 47.28 & 39.11 & 45.20 & 907.10 & 57.30  \\
MetaClusterFastKAN & 5 & -2,2 & 2 & 16 & 48.19 & 45.48 & 47.29 & 611.72 & 57.30  \\

\hline
\\
MetaClusterFastKANConv & 5 & -3,3 & 4 & 256 & 71.16 & 66.46 & 69.58 & 6,077.09 & 428.55  \\
MetaClusterFastKANConv & 5 & -3,3 & 3 & 256 & 73.78 & 70.21 & 72.39 & 4,561.59 & 428.55 \\
MetaClusterFastKANConv & 5 & -3,3 & 2 & 256 & 71.42 & 71.05 & 71.33 & 3,046.08 & 428.55  \\

\hline
\end{tabular}}
\label{tab:embedsizeextended}
\end{table*}

\subsection{Dimensionality and Clusterability Analysis}
\label{app:pca}

\subsubsection{PCA for Lower Dimensional Manifold}
To empirically verify that the meta-learner successfully restricts the KAN activation coefficients to a low-dimensional manifold, we perform Principal Component Analysis (PCA) on the learned edge functions. By examining the explained variance ratio of the principal components, we can quantify the effective dimensionality of the coefficient space. As detailed in Section \ref{sec:manifold}, the meta-learner maps a low-dimensional embedding ($z_i \in \mathbb{R}^{d_{emb}}$) to the full set of KAN weights, constraining them to a task-aligned manifold. For our fully-connected MetaKAN variant on CIFAR-10, we utilize an embedding dimension of $d_{emb}=1$.

\begin{figure}[h]
  \centering
  \begin{subfigure}{0.48\textwidth}
    \centering
    \includegraphics[width=\linewidth]{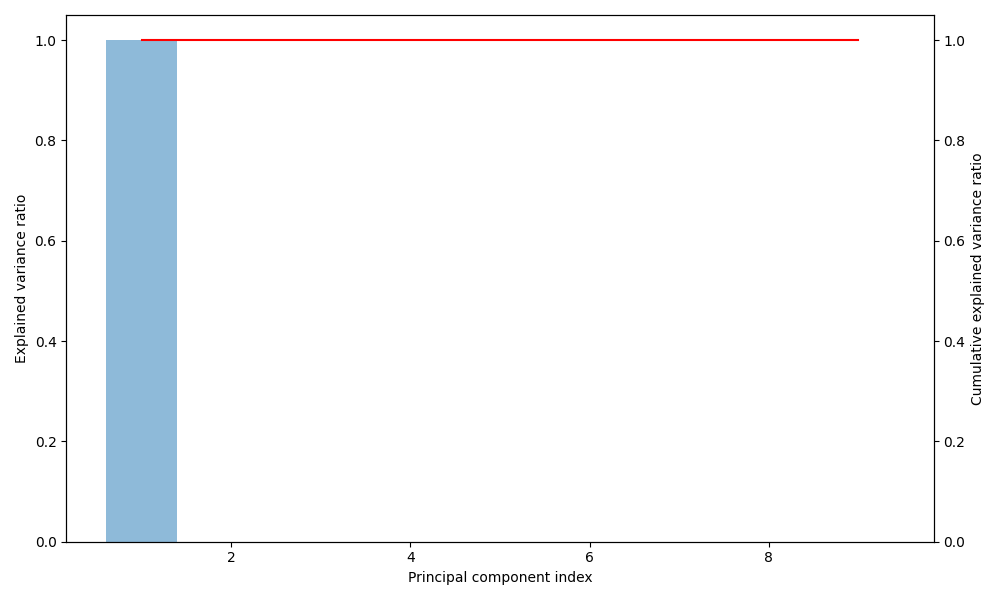}
    \caption{Layer 1 MetaKAN PCA plot.}
    \label{fig:PCAMeta1}
  \end{subfigure}
  \hfill 
  \begin{subfigure}{0.48\textwidth}
    \centering
    \includegraphics[width=\linewidth]{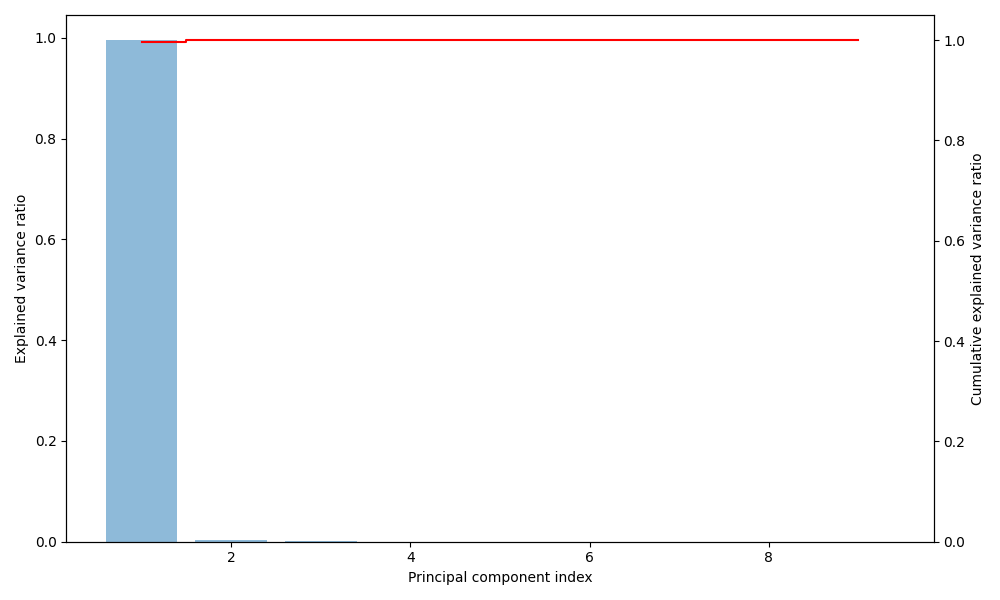}
    \caption{Layer 2 MetaKAN PCA plot.}
    \label{fig:PCAMeta2}
  \end{subfigure}

  \caption{PCA explained variance ratio for the (a) first and (b) second layers of the fully-connected MetaKAN model on CIFAR-10. The single dominant bar demonstrates that the variance is almost entirely explained by the first principal component, confirming the effective restriction of coefficients to a 1D manifold via the $d_{emb}=1$ constraint.}
  \label{fig:PCAMeta}
\end{figure}

Figure \ref{fig:PCAMeta} displays the PCA explained variance ratio for the first and second layers of this fully-connected MetaKAN model. As expected, nearly 100\% of the variance in the generated weights is captured by the very first principal component. This confirms that the meta-learner effectively enforces a one-dimensional manifold, making the resulting high-dimensional coefficient vectors highly compressible and amenable to K-means clustering.

\begin{figure}[h]
  \centering
  \begin{subfigure}{0.48\textwidth}
    \centering
    \includegraphics[width=\linewidth]{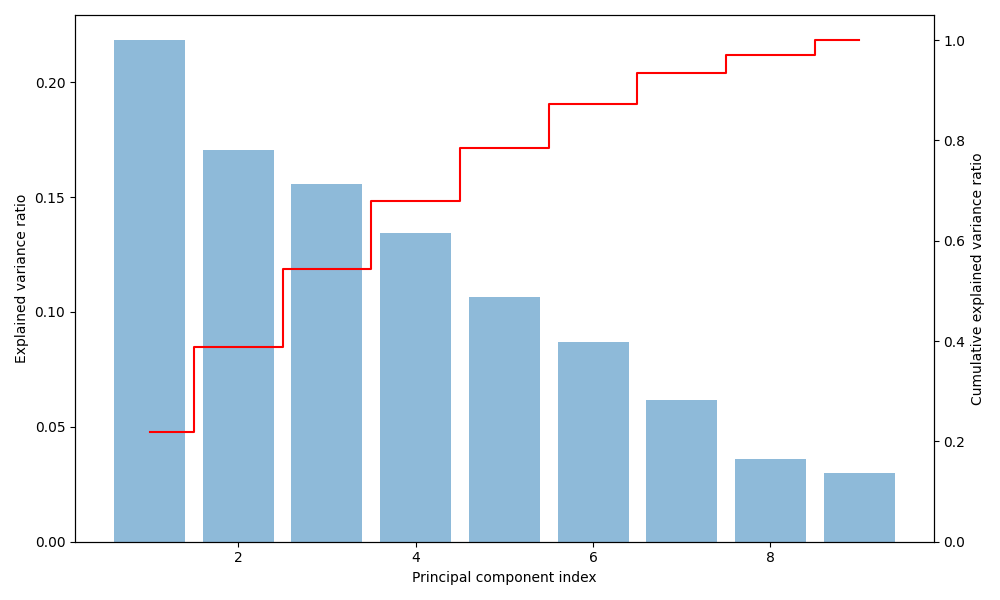}
    \caption{Layer 1 KAN PCA plot.}
    \label{fig:PCAKAN1}
  \end{subfigure}
  \hfill 
  \begin{subfigure}{0.48\textwidth}
    \centering
    \includegraphics[width=\linewidth]{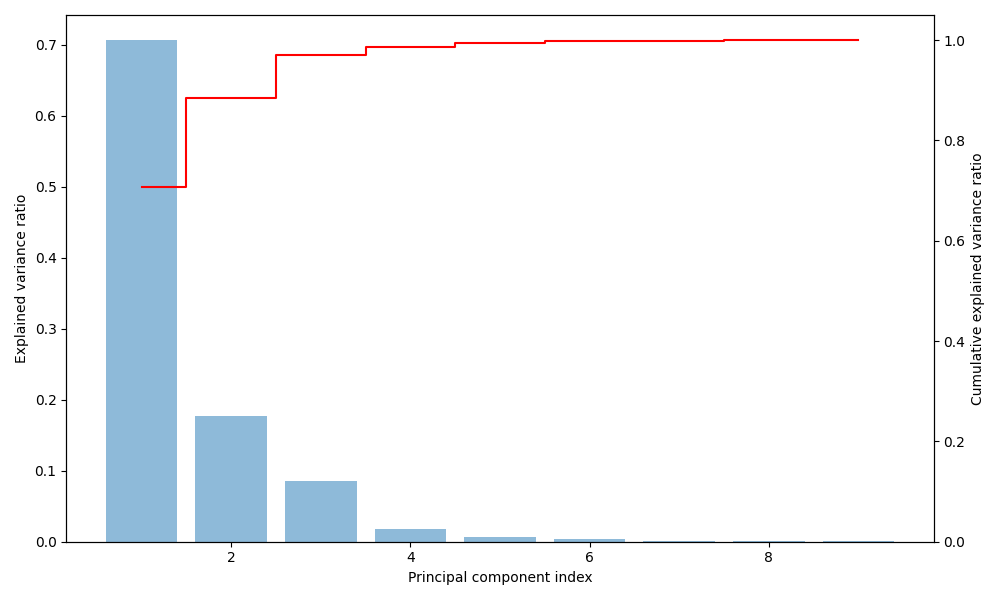}
    \caption{Layer 2 KAN PCA plot.}
    \label{fig:PCAKAN2}
  \end{subfigure}

  \caption{PCA explained variance ratio (blue bars) and cumulative explained variance (red line) for the (a) first and (b) second layers of the baseline fully-connected KAN model on CIFAR-10. The gradual accumulation of variance across multiple components highlights the absence of a distinct low-dimensional manifold when a meta-learner is not utilized.}
  \label{fig:PCAKAN}
\end{figure}

In contrast, Figure \ref{fig:PCAKAN} illustrates the PCA explained variance ratio for the baseline KAN model trained without a meta-learner. Without the structural bottleneck imposed by the low-dimensional embedding, the variance of the learned coefficients is spread across multiple principal components. The cumulative explained variance (shown by the red line) takes several components to approach 1.0. This lack of a coherent low-dimensional structure fundamentally explains why standard clustering methods struggle to achieve high compression rates without significant accuracy degradation in baseline KANs.

\subsubsection{Quantitative Topological Evaluation}
To further confirm that the meta-learner successfully shapes the learned edge coefficients into a more clusterable space, we evaluate standard cluster validation metrics on the first-layer coefficient vectors on CIFAR-10 ($k=16$):

\begin{table}[h]
\centering
\caption{Clustering quality metrics for KAN coefficient vectors on CIFAR-10.}
\label{tab:cluster_metrics}
\begin{scriptsize}
\begin{sc}
\begin{tabular}{lccc}
\toprule
\textbf{Model} & \textbf{Silhouette} ($\uparrow$) & \textbf{Calinski-Harabasz} ($\uparrow$) & \textbf{Davies-Bouldin} ($\downarrow$) \\
\midrule
KAN & $0.08$ & $4{,}678.13$ & $1.82$ \\
MetaClusterKAN  & $\mathbf{0.52}$ & $\mathbf{359{,}394.87}$ & $\mathbf{0.52}$ \\
\bottomrule
\end{tabular}
\end{sc}
\end{scriptsize}
\end{table}

As shown in Table \ref{tab:cluster_metrics}, using the meta-learner increases the Silhouette Coefficient from $0.08$ to $0.52$ and substantially increases the Calinski-Harabasz Index, while reducing the Davies-Bouldin Index from $1.82$ to $0.52$. These results quantitatively support our PCA findings, demonstrating that constraining weights to a lower-dimensional manifold yields tighter, better-separated clusters that make downstream K-means clustering far more effective.

\subsection{Visualizing Clusters}
\begin{figure}[t]
  \centering
  \begin{subfigure}{\textwidth}
    \centering
    \includegraphics[width=0.9\linewidth]{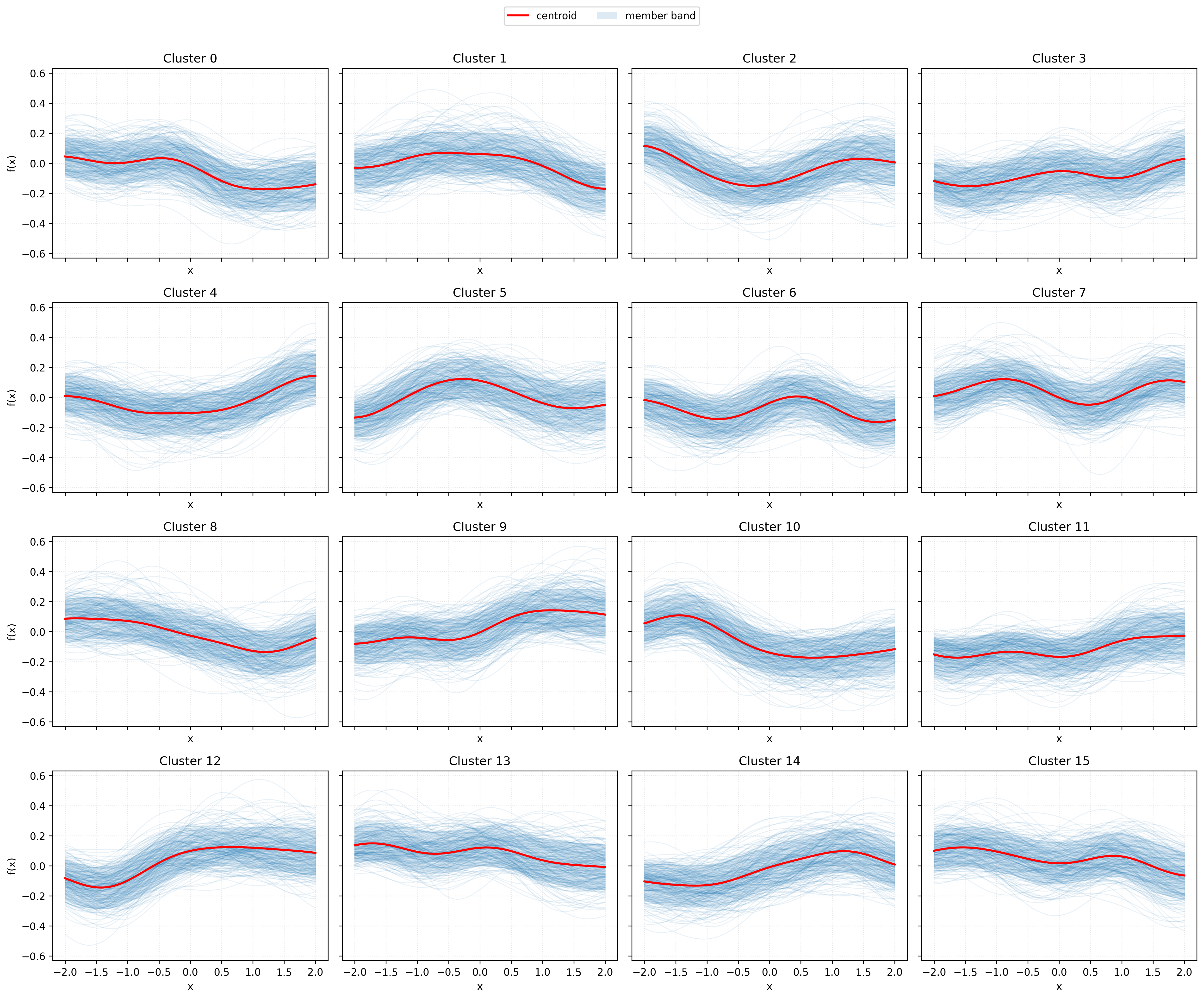}
    \caption{The set of clusters showcasing the edge functions for each cluster on the first layer.}
    \label{fig:metacentroid0}
  \end{subfigure}\hfill
  \begin{subfigure}{\textwidth}
    \centering
    \includegraphics[width=0.9\linewidth]{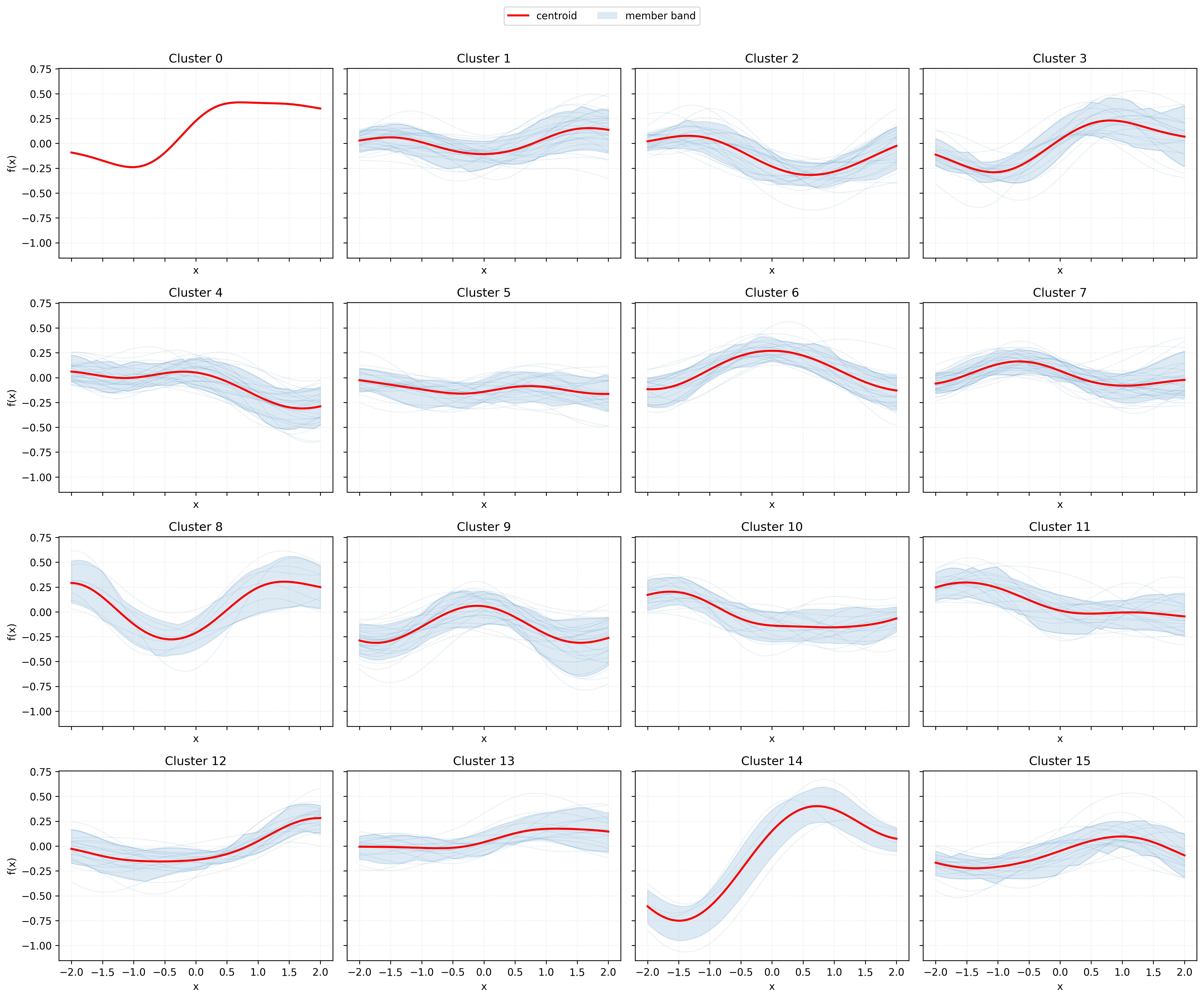}
    \caption{The set of clusters showcasing the edge functions for each cluster on the second layer.}
    \label{fig:metacentroid0}
  \end{subfigure}

  \caption{The edge functions for each cluster (in blue) and the centroid functions (in red) for the (a) first and (b) second layers of the fully-connected FastKAN models on Cifar-10.
  }
  \label{fig:non_metacentroid}
\end{figure}

\begin{figure}[t]
  \centering
  \begin{subfigure}{\textwidth}
    \centering
    \includegraphics[width=0.9\linewidth]{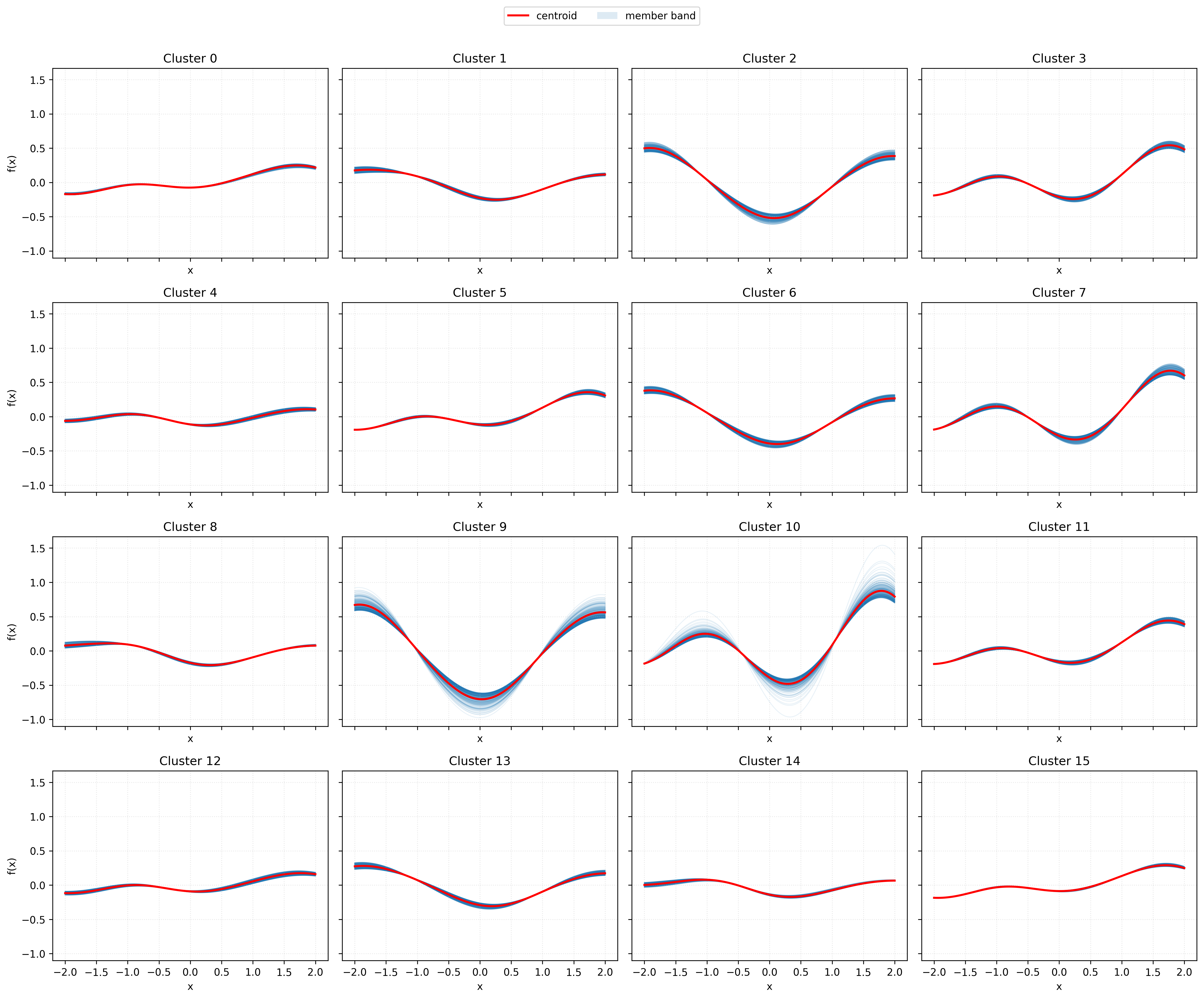}
    \caption{The set of clusters showcasing the edge functions for each cluster on the first layer.}
    \label{fig:metacentroid0}
  \end{subfigure}\hfill
  \begin{subfigure}{\textwidth}
    \centering
    \includegraphics[width=0.9\linewidth]{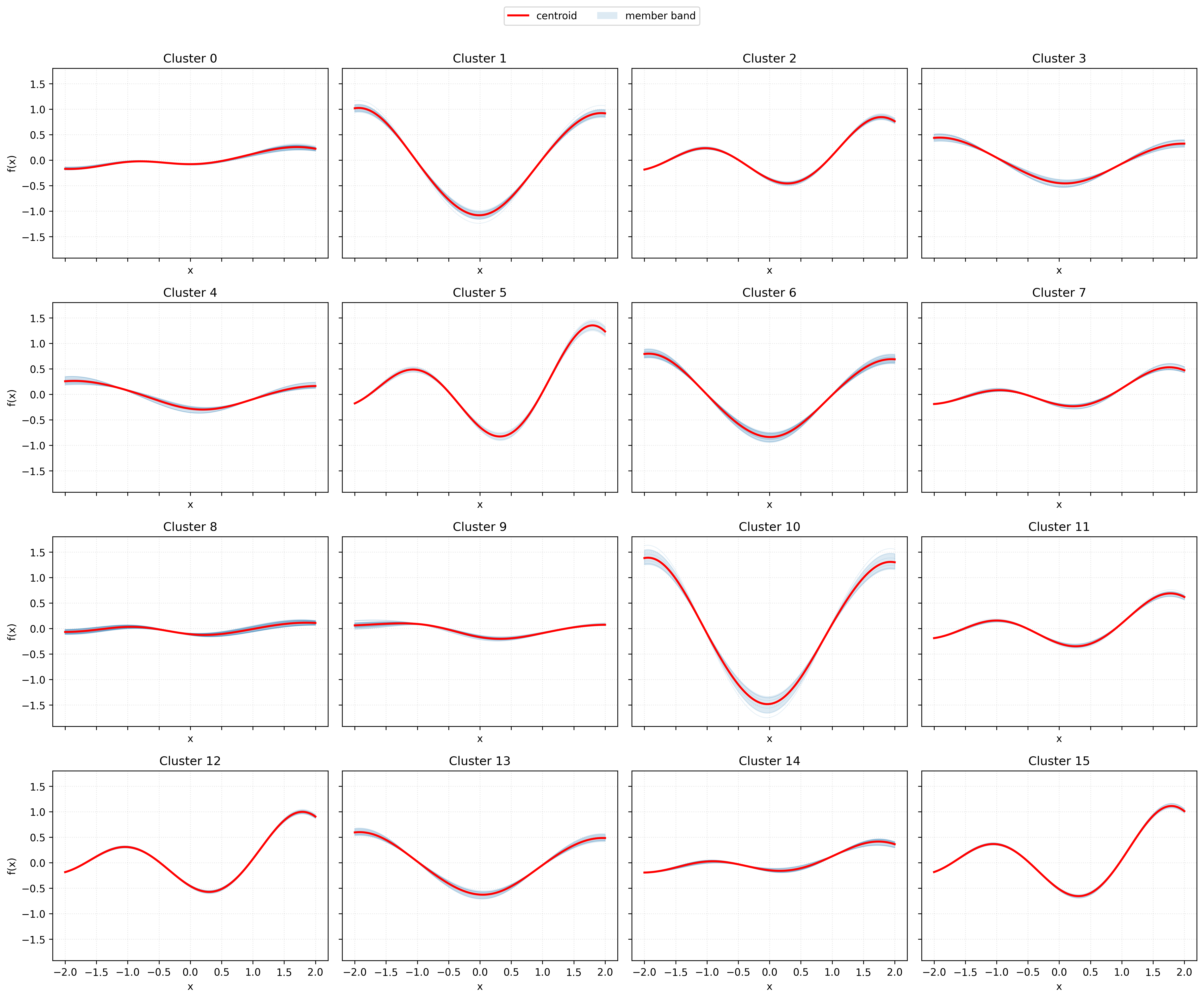}
    \caption{The set of clusters showcasing the edge functions for each cluster on the second layer.}
    \label{fig:metacentroid0}
  \end{subfigure}

  \caption{The edge functions for each cluster (blue) and the centroid functions (red) for the (a) first and (b) second layers of the fully-connected MetaFastKAN models on Cifar-10.
  }
  \label{fig:metacentroid}
\end{figure}

We verify the importance of MetaCluster qualitatively by plotting edge functions and the respective centroids associated with the first layer of FastKAN and MetaFastKAN in Figures \ref{fig:non_metacentroid} and \ref{fig:metacentroid}, respectively.   From observing the functions associated with FastKAN and MetaFastKAN, it is visually clear that the functions associated with MetaFastKAN are much closer to the respective centroid than in FastKAN.  Such a result demonstrates the importance of using meta-learners to find a lower-dimensional functional space before clustering.

\label{app:visualizing}
\FloatBarrier

\end{document}